\documentclass[runningheads]{llncs}

\usepackage{amsmath,amssymb,eurosym,eufrak}
\usepackage{xspace}
\usepackage{subcaption}
\usepackage[utf8]{inputenc}
\usepackage{enumitem}
\usepackage{stmaryrd}
\usepackage{booktabs} 
\usepackage{hyperref}
\usepackage{fixfoot}
\usepackage{semantic}
\usepackage{bookmark}
\usepackage{adjustbox}
\usepackage{wrapfig}
\usepackage[linesnumbered,ruled,noend]{algorithm2e}
\usepackage{cleveref}
\usepackage{multirow}
\usepackage[framemethod=tikz]{mdframed}
\usepackage{float}

\usepackage{tikz}
\usepackage{pgfplots}
\usepackage{pgfplotstable}
\pgfplotsset{compat=1.16}
\usetikzlibrary{shapes,shapes.geometric,arrows,arrows.meta,fit,calc,positioning,automata}
\usetikzlibrary{pgfplots.statistics}
\usepgfplotslibrary{fillbetween} 

\definecolor{darkgreen}{rgb}{0,0.6,0}

\renewcommand{\subsubsection}[1]{\leavevmode\unskip\\\noindent\textbf{#1.}}

\newcommand{\cM}{\ensuremath{\mathcal{M}}\xspace}

\newcommand{\cU}{\ensuremath{\mathcal{U}}\xspace}

\newcommand{\Act}{\mathit{Act}}
\newcommand{\Post}{\mathit{Post}}

\DeclareMathOperator*{\argmax}{arg\,max}
\DeclareMathOperator*{\argmin}{arg\,min}

\makeatletter
\newcommand{\oset}[3][-.2ex]{%
  \mathrel{\mathop{#3}\limits^{
    \vbox to#1{\kern-2\ex@
    \hbox{$\scriptstyle#2$}\vss}}}}
\makeatother

\newcommand{\hframe}[1]{\begin{mdframed}[backgroundcolor=lightgray!30,%
	linecolor=lightgray!30,%
	innertopmargin = 3pt,%
	innerleftmargin = 3pt,%
	innerrightmargin = 3pt]
#1
\end{mdframed}}

\newcommand{\Nat}{\mathbb{N}}

\newcommand{\Real}{\mathbb{R}}

\newcommand{\ra}{\ensuremath{\rightarrow}\xspace}

\newcommand{\pa}{\ensuremath{\rightharpoonup}\xspace}

\newcommand{\Nt}{N_t}

\newcommand{\Dist}{\mathit{Dist}}
\newcommand{\Intv}{\mathit{Intv}}
\newcommand{\Exp}{\mathbb{E}}
\newcommand{\Prob}{\mathbb{P}}
\newcommand{\Runs}{\mathit{Runs}}

\newcommand{\python}{\textsc{Python}}

\newcommand{\racetrack}{\textsc{RaceTrack}}

\SetKwFor{RepTimes}{repeat}{times}{end}

\newcommand{\thetitle}{Strategy Synthesis in Markov Decision Processes Under Limited Sampling Access}

\begin{document}
\title{\thetitle%
 \thanks{The authors are supported by the DFG through
	the Cluster of Excellence EXC 2050/1 (CeTI, project ID 390696704, as part of Germany's 
 	Excellence Strategy) and the TRR 248 (see \url{https://perspicuous-computing.science}, project ID 389792660).}
    }
\titlerunning{MDP Strategy Synthesis Under Limited Sampling Access}

\author{Christel Baier\inst{1,2}
\and Clemens Dubslaff\inst{4,1}
\and Patrick Wienh\"oft\inst{1,2}
\and Stefan J. Kiebel\inst{1,3}
}

\institute{
	Centre for Tactile Internet with Human-in-the-Loop (CeTI)\\
  \and
  Department of Computer Science, Technische Universit\"at Dresden, Dresden, Germany\\
	\and
	Department of Psychology, Technische Universit\"at Dresden, Dresden, Germany\\
  \email{\{christel.baier,patrick.wienhoeft,stefan.kiebel\}@tu-dresden.de}  
	\and
	Eindhoven University of Technology, Eindhoven, The Netherlands
	\email{c.dubslaff@tue.nl}
}

\maketitle

\begin{abstract}
	A central task in control theory, artificial intelligence, and formal 
	methods is to synthesize reward-maximizing strategies for agents that operate in 
	partially unknown environments. 
	In environments modeled by \emph{gray-box} Markov decision processes (MDPs), 
	the impact of the agents' actions are known in terms of successor states but 
	not the stochastics involved.
	In this paper, we devise a strategy synthesis algorithm for gray-box MDPs via
	reinforcement learning that utilizes \emph{interval MDPs} as internal model.
	To compete with limited sampling access in reinforcement learning, 
	we incorporate two novel concepts into our algorithm,
	focusing on rapid and successful learning rather than on stochastic guarantees and optimality:
	\emph{lower confidence bound} exploration 
	reinforces variants of already learned practical strategies 
	and \emph{action scoping} 
	reduces the learning action space to promising actions.
	We illustrate benefits of our algorithms by means of a prototypical implementation 
	applied on examples from the AI and formal methods communities.
\end{abstract}


\section{Introduction}\label{sec:introduction}
Many machine learning methods take inspiration from the inner-workings of the human 
brain or human behavior~\cite{Mitchell1997}. For instance, learning based
on neural networks mimics the human brain at a structural level by explicitly modeling
its neurons and their activation. Taking a more high-level
view, \emph{reinforcement learning (RL)}~\cite{SutBar18} formalizes human
learning behavior by reinforcing actions that are repeatedly associated 
with successful task solving~\cite{And00}.
The usual application of RL is to learn reward-optimizing strategies
in environments modeled as \emph{Markov decision processes (MDPs)}~\cite{Puterman} 
where the agent has only partial knowledge and 
learns based on guided exploration through sample runs.
Existing RL approaches prioritize stochastic guarantees and convergence 
to a globally optimal strategy, leading to slow learning performance and infeasibility
for small sample sizes~\cite{SutBar18}.
In contrast, human decision making can compete with limited sampling access, not focusing
on strict optimality but on efficiency. 
The more urgent a task and the less time available for its solving, 
the more humans tend to exploit previously learned strategies -- 
possibly sacrificing optimality but increasing the chance of finishing the task in time~\cite{Schwoebel21}. 
In the extremal case, humans rely on \emph{habits}~\cite{WooRun16}, i.e., 
sequences of actions that, once triggered, are executed mostly independent from 
reasoning about the actual task~\cite{BaiRivDub21}. 
Habits avoid further costly exploration during learning by restricting 
the action space.

In this paper, we take inspiration from humans' ability to reason efficiently 
with few explorations, shaping novel RL algorithms that rapidly synthesize ``good'' strategies.
Specifically, our learning task amounts to an agent being able to determine a strategy with high expected 
accumulated reward until reaching a goal, given a limited number of samples.
We consider the setting where the environment is modelled as a \emph{contracting MDP}, i.e., 
goal states are almost surely reached under all strategies,
on which the agent has a \emph{gray-box} view, i.e., 
knows the reward structures and the topology but not the exact probabilities~\cite{AshKreWei19}.
We tackle this task of sample-bounded learning towards nearly-optimal strategies
by introducing two new concepts: \emph{lower confidence bound (LCB) sampling} and \emph{action scoping}.
Classical reward-based sampling in RL is based on upper confidence bounds (UCB)~\cite{Amin21}, 
balancing the \emph{exploration-exploitation dilemma}~\cite{SutBar18}. 
In contrast, our LCB sampling method favors situations already shown viable during the learning process. 
Hence, exploration is limited when there are no good reasons for leaving 
well-known paths, similar to what humans do with habitual sequences of actions~\cite{WooRun16}.
The second learning component is \emph{action scoping}, 
restraining exploration actions when shown to be suboptimal in past samples.
Scoping is parametrized to tune the degree of exploration and balance between fast strategy synthesis
or increasing the chance of learning optimal strategies.

To implement our novel concepts, we provide technical contributions by presenting an RL algorithm
on contracting gray-box MDPs with arbitrary rewards and various sampling methods.
The learning algorithm is a sample-based approach that generates an \emph{interval MDP (IMDP)}
to approximate the environment and whose intervals are iteratively refined.
While methods for analyzing IMDPs have already been considered in the literature~\cite{Givan00,Wu08}, 
and IMDPs have been used in the context of RL algorithms \cite{SuiSimJan22}, we 
provide a new connection of their use in PAC RL algorithms.
We devise our human-inspired RL algorithms, including LCB and action scoping, 
by modeling knowledge of the agent as IMDP
using concepts from \emph{model-based interval estimation (MBIE)}~\cite{Strehl08} 
and \emph{probably almost correct (PAC) statistical model checking (SMC)}~\cite{AshKreWei19}.
We show that our algorithms on IMDPs are PAC for UCB and LCB sampling,
i.e., the probability 
of a suboptimal strategy can be quantified by an arbitrarily small error tolerance.
This, however, cannot be guaranteed in the case of action scoping.
Towards an evaluation of LCB and action scoping, we implemented our algorithms 
in a prototypical tool~\cite{artifact}.
By means of several experimental studies from the RL and formal-methods community, e.g., 
on multi-armed bandits~\cite{Web92} and \racetrack\ \cite{BarBraSin95}, we show that LCB and action scoping 
foster fast strategy synthesis, providing better strategies after fewer sample runs than RL-style
PAC-SMC methods.
We discuss the impact of scoping parameters and related
heuristics, as well as combinations of sampling strategies.
In summary, our contributions are:
\begin{itemize}
	\item A (model-based) RL algorithm for contracting gray-box MDPs with integer rewards
		relying on IMDP and sampling strategy refinements
	(see \Cref{sec:algorithms})
	\item Instances of this RL algorithm subject to
				lower and upper confidence bound sampling and
			tunable action scoping (see \Cref{sec:habits}).
	\item A prototypical implementation of our RL algorithms and an
	evaluation 
		in examples from both the RL and formal-methods communities.
\end{itemize}

\subsubsection{Related work}
SMC~\cite{Legay2019} for unbounded temporal properties in stochastic systems is most related to 
our setting,
establishing algorithms also in gray-box settings~\cite{YouClaZul10,HeJenBas10}.
Given a lower bound on transition probabilities, SMC algorithms have been presented 
for Markov chains~\cite{DacHenKre16}, MDPs, and even stochastic games~\cite{AshKreWei19}.
Recent SMC algorithms for MDPs also include learning~\cite{BraChaChm14,AshKreWei19}
but only for reachability problems.
IMDPs have been investigated outside of the RL context in formal 
verification~\cite{SenVisAgh06,ChaSenHen08} for $\omega$-regular properties,
for positive rewards in contracting models~\cite{Wu08} by an extension of the well-known 
value-iteration algorithm~\cite{SutBar18}, and
in the performance-evaluation community in the discounted setting \cite{Givan00}.
In particular, the RL algorithms we present in this paper use an adaptation of the latter algorithm
without discounting as a subroutine to successively tighten bounds on the maximal expected accumulated rewards.
More recently, algorithms with convergence guarantees for reachability
objectives in (interval) MDPs have been presented \cite{BeiKleLeu17,HadMon18}.
\emph{Interval estimation} for RL has been introduced by Kaelbling~\cite{Kae93}
towards Q-learning~\cite{WatDay92} and extended to model-based approaches~\cite{Wie98}
such as MBIE~\cite{Strehl04} 
and
the UCRL2 algorithm \cite{JakOrtAue10} using an error tolerance based on the $L_1$-norm
opposed to the $L_\infty$-norm employed in interval MDPs.
In contrast, the \emph{linearly updating intervals} \cite{SuiSimJan22} algorithm utilizes
IMDPs but uses potentially unsafe intervals and focuses on learning on changing environments.
Besides UCB sampling, the exploration-exploitation dilemma in reward-based learning 
has also be addressed with exploration bonuses~\cite{Kae93,KaeLitMoo96,Sut91,ThrMol92},
performing well when applied to MBIE~\cite{Ish02,Strehl08}
or in other RL methods such as E$^3$~\cite{KeaSin02} and R$_\mathrm{max}$~\cite{Rmax}.

\subsubsection{Supplements}
The appendix contains proofs and full experimental evaluations.
Our implementation and data sets to reproduce the experiments of this paper
is available~\cite{artifact}.

\section{Preliminaries}\label{sec:preliminaries}

A \emph{distribution} over a finite set $X$ is a function $\mu\colon X\rightarrow [0,1]$
where $\sum_{x\in X} \mu(x) = 1$. The set of distributions over $X$ is denoted by
$\Dist(X)$.

\subsubsection{Markov decision processes (MDPs)} 
An MDP is a tuple $\cM = (S,A,\imath,G,R,T)$ where 
$S$, $A$, and $G \subseteq S$ are finite sets of states, actions, and goal states, respectively,
$\imath \in S$ is an initial state, $R \colon S \rightarrow \Real$ is a reward function,
and $T\colon S{\times}A \pa \Dist(S)$ is a partial transition probability function.
For state $s\in S$ and action $a\in A$ we say that $a$ is \emph{enabled} in $s$ if
$T(s,a)$ is defined. We assume the set $\Act(s)$ of all enabled actions to be empty
in goal states $s\in G$ and non-empty in all other states.
For $(s,a,s')\in S{\times}A{\times}S$ we define $T(s,a,s')=T(s,a)(s')$ 
if $T(s,a)$ is defined and $T(s,a,s')=0$ otherwise.
The \emph{successors} of $s$ via $a$ are denoted by $\Post(s,a) = \{ s' \mid T(s,a,s')>0 \}$.
A \emph{run} of $\cM$ is a sequence $\pi = s_0a_0s_1a_1\dots s_n$
where $s_0=\imath$, $s_n \in S$, $(s_i,a_i)\in (S{\setminus}G)\times A$, 
and $s_{i+1}\in\Post(s_i,a_i)$ for $i= 0,\ldots,n{-}1$.
The set of all runs in $\cM$ is denoted by $\Runs(\cM)$.
The \emph{accumulated reward} of $\pi$ is defined by $R(\pi) = \sum_{i=0}^{n-1} R(s_i)$.

An \emph{interval MDP (IMDP)} is a tuple
$\cU=(S,A,\imath,G,R,\hat{T})$ where $S,A,\imath,G$, and $R$ are as 
for MDPs, and $\hat{T} \colon S {\times} A \pa \Intv(S)$ is an
\emph{interval transition function}. Here, $\Intv(S)$ denotes the set of interval functions
$\nu\colon S \ra \{ [a,b] \mid 0 < a \leq b \leq 1 \} \cup \{ [0,0] \}$ over $S$.
Note that $\Dist(S)\subseteq\Intv(S)$, i.e., every distribution over $S$ is also
an interval function. A distribution $\mu\in\Dist(S)$ is an \emph{instantiation} of 
$\nu\in\Intv(S)$ if $\mu(s)\in\nu(s)$ for all $s\in S$. 
We again say $a$ is enabled in $s$ if $\hat{T}(s,a)$ is defined and denote
the set of enabled actions in $s$ as $\Act(s)$, assumed to be non-empty for all $s\in (S\setminus G)$.
For each $s\in S$ and $a\in \Act(s)$ we denote by $T_s^a$ the set of 
all instantiations $t_s^a$ of $\hat{T}(s,a)$ and define $\Post(s,a)=\{s' \mid \underline{T}(s,a,s')>0\}$.
The MDP $\cM$ is an \emph{instantiation} of $\cU$ if 
$T(s,a)\in T_s^a$ for all $s\in S$, $a\in A$.
We denote by $[\cU]$ the set of all instantiations of $\cU$.
Note that as all instantiations of an IMDP $\cU$ share the same topology, 
the set of runs $\Runs(\cM)$ is the same for all instantiations $\cM\in[\cU]$.

The semantics of the MDP $\cM$ is given 
through \emph{strategies}, i.e., mappings $\sigma\colon S
\rightarrow \Dist(A)$ where $\sigma(s)(a)=0$ for all $a\not\in\Act(s)$. 
We call a run $\pi=s_0a_0s_1a_1\dots s_n$ 
in $\cM$ a \emph{$\sigma$-run} if $\sigma(s_i)(a_i)>0$ for all $i=0,\ldots,n{-}1$. 
The probability of $\pi$ is defined as $\Pr^\sigma(\pi) = \prod_{i=0}^{n-1} \sigma(s_i)(a_i)\cdot T(s_i,a_i,s_{i+1})$ if $\pi$ is a $\sigma$-run and $\Pr^\sigma(\pi)=0$ otherwise.
The probability of some $B\subseteq\Runs(\cM)$ w.r.t. strategy $\sigma$ is defined
by $\Pr^\sigma(B) = \sum_{\pi\in B} \Pr^\sigma(\pi)$.
If $\Pr^\sigma(B)=1$, then the \emph{expected (accumulated) reward} is defined
as $\Exp^\sigma(B) = \sum_{\pi\in B} \Pr^\sigma(\pi)\cdot R(\pi)$.
We call $\cM$ \emph{contracting} \cite{KalLN20} if $\Pr^\sigma(\lozenge G) = 1$ for all strategies $\sigma$, i.e., 
a goal state is almost surely reached for any strategy. 
The semantics of an IMDP $\cU$ is the set of its instantiations $[\cU]$. 
An IMDP $\cU$ is \emph{contracting} iff all MDPs in $[\cU]$ are contracting. 
Note that for IMDPs there is also a notion of an operational semantics that lifts strategies
to instantiations of transitions \cite{SenVisAgh06}. While different in its nature, our contributions 
of this paper can be easily extended also to the latter semantics.

\subsubsection{Value and quality functions}
A \emph{value function} $V_{\cM} \colon S \rightarrow \Real$ of MDP $\cM$ is the solution of the 
\emph{Bellman equations}~\cite{BerTsi91} given by $V_\cM(s)=R(s)$ for $s\in G$ and
\begin{center}
$V(s)\ =\
	R(s) + \max_{a\in \Act(s)} \sum_{s' \in S} V_\cM(s')\cdot T(s,a,s')$ \quad for \enskip $s\not\in G$.
\end{center}
The \emph{quality} $Q_\cM\colon S\times A \pa \Real$ of $\cM$ is defined for all $s\in S$ and $a\in \Act(s)$ by
\begin{center}$
	Q_\cM(s,a)\ =\ R(s)\ + \sum\nolimits_{s' \in \Post(s,a)} V_\cM(s')\cdot T(s,a,s')
$\end{center}
Intuitively, the quality represents the value of choosing 
an action $a$ in state $s$ continuing with a reward-maximizing strategy.
For an IMDP $\cU$, the value function differs between instantiations, leading to
Bellman equations

\begin{center}
	$\underline{V}_\cU(s)\ =\  \min_{\cM\in [\cU]}{V_\cM(s)} \quad\quad
\overline{V}_\cU(s) =\max_{\cM\in [\cU]}{V_\cM(s)}$
\end{center}
for the lower and upper bounds on possible instantiations, respectively. 
These value functions are naturally lifted to quality functions for IMDPs.
We omit subscript $\cM$ or $\cU$ if clear from the context.
Further, we define the \emph{pessimistically optimal strategy}
$\underline{\sigma}$ for all $s\in (S\setminus G)$ as
$\underline{\sigma}(s)=\argmax_{a\in\Act(s)}\underline{Q}(s,a)$ and similarly the
\emph{optimistically optimal strategy} as
$\overline{\sigma}(s)=\argmax_{a\in\Act(s)}\overline{Q}(s,a)$.

\begin{figure}[t]
	\resizebox{\textwidth}{!}{\input{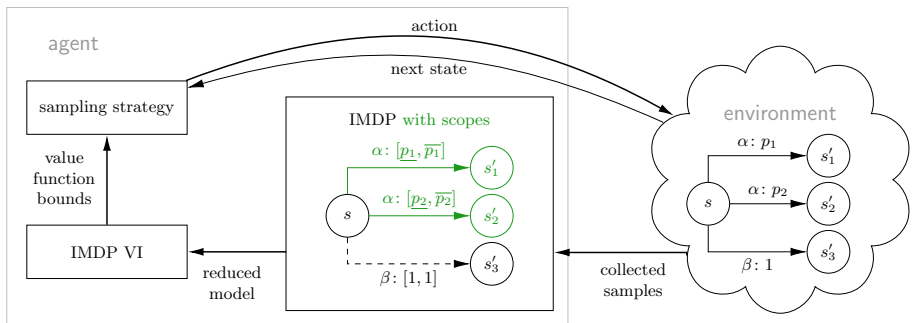}}
	\caption{\label{fig:workflow}Schema of reinforcement learning in gray-box MDPs}
\end{figure}


\section{Interval MDP reinforcement learning}\label{sec:algorithms}
In this section, we establish an RL algorithm for contracting gray-box MDP
that generates a white-box IMDP and successively shrinks the transition probability intervals 
of the IMDP while updating the sampling strategy.
Let $\cM=(S,A,\imath,G,R,T)$ be a contracting MDP as above, serving as environmental model.
With RL in a \emph{gray-box} setting, the agent's objective is to determine 
reward-maximizing strategies knowing
all components of $\cM$ except transition probabilities $T$.
We further make the common assumption~\cite{AshKreWei19,BeiKleLeu17,DacHenKrePet2017}
that there is a known constant $p_{min}$ 
that is a lower bound on the minimal transition probability, i.e.,
$p_{min} \leq \min \{ T(s,a,s') \mid T(s,a,s')>0\}$.

To learn strategies in $\cM$, samples are generated according to a \emph{sampling strategy}
determining the next action an agent performs in each state.
\Cref{fig:workflow} shows the overall schema of the algorithm, which
runs in \emph{episodes}, i.e., batches of samples. 
The sampling strategy is updated after each episode by refining an internal IMDP model 
based on the sample runs and an IMDP value iteration.

\subsection{Generating IMDPs from sampled gray-box MDPs}\label{subsec:mb}
Let $\#(s,a,s')$ denote the number of times the transition $(s,a,s')$
occurred in samples
thus far and let $\#(s,a)=\sum_{s'\in\Post(s,a)} \#(s,a,s')$.
The goal of each episode is to approximate $\cM$ by an IMDP 
$\cU=(S,A,\imath,G,R,\hat{T})$ that is \emph{($\mathit{1-}\delta$)-correct},
i.e., the probability of $\cM$ being an instantiation of $\cU$ is at
least $1{-}\delta$ for a given error tolerance $\delta\in\Real$. Formally
$\prod_{(s,a)\in S\times A} \Prob\big(T(s,a)\in\hat{T}(s,a)\big) \geqslant 1{-}\delta$~\cite{AshKreWei19},
where $\mathbb{P}$ refers to the probabilistic behaviour of the algorithm
due to sampling the gray-box MDP.
The idea towards $(1{-}\delta)$-correct IMDPs is to distribute the error 
tolerance $\delta$ over transitions by defining a 
\emph{transition error tolerance} $\eta\in\Real$.
Given a state $s\in S$, an action $s\in\Act(s)$ and a successor $s'\in\Post(s,a)$,
we define the interval transition probability function 
$\hat{T}_\eta\colon S{\times}A\ra\Intv(S)$ as
\begin{center}
	$\hat{T}_\eta(s,a,s') = 
	\left[\frac{\#(s,a,s')}{\#(s,a)}-c(s,a,\eta),
	\frac{\#(s,a,s')}{\#(s,a)}+c(s,a,\eta)\right] 
	\cap \left[p_{min},1\right].$
\end{center}
where $c(s,a,\eta)=\sqrt{\frac{\log\eta/2}{-2\#(s,a)}}$.
Hoeffding's inequality \cite{Hoeffding} then yields $T_\eta(s,a)\in\hat{T}(s,a)$ with probability
at least $1{-}\eta$. 
To instantiate an environment approximation, we distribute the error tolerance $\delta$
\emph{uniformly}, i.e., to define $\hat{T}_\eta$ and thus obtain $\cU$ 
we set $\eta = \delta / \Nt$ where $\Nt$ 
is the overall number of probabilistic transitions in $\cM$, i.e.,
$\Nt = \lvert\{ (s,a,s') \mid s'\in\Post(s,a) \text{ and } \lvert\Post(s,a)\rvert>1 \}\rvert$.
Note that $\Nt$ only depends on the topology of the MDP and is thus known
in a gray-box setting.

\begin{algorithm}[t]
	\SetAlgoLined
	\DontPrintSemicolon
	\SetKwInOut{Input}{Input}\SetKwInOut{Output}{Output}
	\Input{gray-box MDP $\cM=(S,A,\imath,G,R,\cdot)$, error tolerance $\delta$,
	$K,N\in\Nat$
	}
	\Output{pessimistically and optimistically optimal strategies $\underline{\sigma}$ and $\overline{\sigma}$, 
		value function bounds $\underline{V}$ and $\overline{V}$}
	\BlankLine
	\ForAll{$(s,a)\in S\times A$}{
		$\sigma(s)(a) := 1 / \lvert \Act(s) \rvert$\hfill\tcp{initialize}
		\lForAll{$s'\in \Post(s,a)$}{
			$\hat{T}(s,a,s') := [p_{min},1]$
		}
	}
	$\cU := (S,A,\imath,G,R,\hat{T})$\;
	\BlankLine
	\ForAll{$k\in \{1,\dots,K\}$\label{l:episode}}{
		\lForAll{$n\in \{1,\dots,N\}$}{	SAMPLE($\cM,\sigma$)\hfill\tcp*[f]{sample runs}\label{l:sample}}
		$\cU :=$ UPDATE\_PROB\_INTERVALS($\cU,\delta$)\label{alg:update}\hfill\tcp{build IMDP model}
		$(\underline{V}, \overline{V}) :=$ COMPUTE\_BOUNDS($\cU, k$)\label{l:bounds}\hfill\tcp{IMDP value iteration}
		$\sigma :=$ UPDATE\_STRATEGY($\cU, \underline{V},\overline{V})$\label{l:policy}\hfill\tcp{compute sampling strategy}
	}
	\ForAll{$s\in(S\setminus G)$}{
		$\big(\underline{\sigma}(s), \overline{\sigma}(s)\big):=\big(\argmax_{a\in\Act(s)}\underline{Q}(s,a),
		\argmax_{a\in\Act(s)}\overline{Q}(s,a)\big)$\;
	}
	\Return $\underline{\sigma},\underline{V}, \overline{\sigma}, \overline{V}$\;
	\caption{IMDP\_RL($\cM,\delta,K,N$)\label{alg:imdp-rl}}
\end{algorithm}

\subsubsection{Value iteration on environment approximations}
We rely on value iteration for IMDPs~\cite{Givan00,Wu08}
to solve the interval Bellman equations for all possible instantiations
of our environment approximation IMDP $\cU$. 
Standard value iteration for IMDPs does not exhibit a stopping criterion
to guarantee soundness of the results. For soundness,
we extend interval value iteration~\cite{BeiKleLeu17,HadMon18}
with a conservative initialization bound for the value function.
For technical details of the value iteration on IMDPs we refer to the appendix.

\subsection{IMDP-based PAC Reinforcement learning}
Piecing together the parts discussed so far, we obtain an IMDP-based RL algorithm
sketched in \Cref{alg:imdp-rl} (cf. also \Cref{fig:workflow}),
comprising $K$ episodes with $N$ sample runs each,
updating the model and performing a value iteration (see \Cref{l:episode} and 
\Cref{l:sample}, respectively).  Both $K$ and $N$ can be seen as parameters
limiting the sampling access of the agent.
The function \textrm{SAMPLE} in \Cref{l:sample} interacts with the environment $\cM$ and chooses either 
a yet not sampled action, or samples an action according to $\sigma$. 
A run ends when entering a goal state $s\in G$, or upon reaching a length of $\lvert S \rvert$.
The latter is to prevent runs from acquiring a large number of samples by simply staying inside a
cycle for as long as possible.
In \Cref{alg:update}, the subroutine \textrm{UPDATE\_PROB\_INTERVALS}
incorporates the fresh gathered samples from the environment into the internal
IMDP representation as outlined in \Cref{subsec:mb}, updating transition probability intervals.
The IMDP value iteration \textrm{COMPUTE\_BOUNDS} in \Cref{l:bounds} yields
new upper and lower value functions bounds.
The number of value iteration steps is $k\cdot\lvert S \rvert$, i.e., increases with each episode to guarantee that
the value function is computed with arbitrary precision for a large number of episodes,
also known as \emph{bounded value iteration}~\cite{AshKreWei19,BraChaChm14}.
The computed bounds are then used in \textrm{UPDATE\_STRATEGY} in \Cref{l:policy}
to update the sampling strategy for the next episode.
The environment approximation $\cU$ can be achieved following several
strategies according to which samples are generated \cite{Amin21}.
A strategy that is widely used in SMC~\cite{AshKreWei19} or
tabular RL~\cite{Auer04,JakOrtAue10,Strehl04,Strehl08} is
\emph{upper confidence bound (UCB)} sampling.
The UCB strategy samples those actions $a$ in state $s$
that have highest upper bound on the quality $\overline{Q}(s,a)$,
resolving the well-known exploration-exploitation dilemma in RL.
This principle is also known as ``optimism in the face of uncertainty'' (OFU),
referring to UCB allocating uncertain probability mass to the best possible outcome~\cite{Amin21}.
In our framework, standard UCB sampling will serve as the baseline approach.
Lastly, we compute and return pessimistic and optimistic strategies
along with their value function bounds, before returning them.
\begin{theorem}\label{thm:guarantees}
Let $V^{*}$ be the solution to the Bellman equations of a given MDP $\cM$. 
Then for all $\delta\in\ ]0,1[$ and $K,N\in\Nat$
the value function bounds $\underline{V}$ and $\overline{V}$ returned by 
\textup{IMDP\_RL($\cM,\delta,K,N$)} as of \Cref{alg:imdp-rl} contain $V^{*}$ 
with probability at least $1-\delta$, 
i.e., $\Prob\left(\underline{V}(s) \leqslant V^{*}(s) \leqslant \overline{V}(s)\right)\geqslant 1-\delta$
for all $s\in S$. 
\end{theorem}
The proof of this theorem is provided in the appendix, essentially showing that (1)
$\cM\in [\cU]$ with probability at least $1-\delta$ and (2) for each state $s$
the solution to the interval Bellman equation of $\cU$ is a subinterval 
of the computed $[\underline{V}(s), \overline{V}(s)]$.


\section{Learning under limited sampling access}\label{sec:habits}
Previous work has shown that RL algorithms utilizing the OFU principle converge towards
an optimal solution and are also able to perform well in practice~\cite{SutBar18}.
However, they are known to converge rather slowly, requiring lots 
of sampling data and training time.
In this section, we use our IMDP-RL algorithm presented in \Cref{alg:imdp-rl}
in a setting where sampling access is limited, i.e., 
the parameters $K$ and $N$ are fixed.
Then, the OFU principle might be not suitable anymore,
as the strategy is learnt under an optimistic view for
increasing confidence in the actions' impacts, which 
requires lots of samples for every action. 
We propose to focus on finding ``good'' strategies 
within the bounded number samples rather than on guaranteed convergence 
to an optimal strategy. Specifically,
we present two complementary methods to reduce the action spaces during sampling:
\emph{lower confidence bound sampling} and \emph{action scoping}.
Both methods are parametrizable and thus can be adapted to the model size as well 
as the bound imposed on the number of samples. 

\subsection{Lower confidence bound sampling}
As new sampling strategy incorporated in \Cref{l:policy} of \Cref{alg:imdp-rl}, 
we propose to choose an action $a$ in a state $s$ 
if it has the highest \emph{lower bound} $\underline{Q}(s,a)$ instead of the highest
\emph{upper bound} as within UCB sampling.
While then the agent still naturally chooses actions that were already sampled often with high
rewards, this avoids further exploring actions with high transition uncertainty.
However, such a \emph{lower confidence bound (LCB)} sampling
might result in performing exploitations only. Hence, we include
an $\epsilon$-greedy strategy~\cite{SutBar18}
into LCB sampling: In each step, with probability $1{-}\epsilon$ the action with the 
highest LCB is sampled and with probability $\epsilon$ a random action is chosen.
In the following, we identify LCB sampling with a degrading $\epsilon$-greedy LCB strategy.
Note that also any other exploration strategies, such as sampling with decaying $\epsilon$ or
\emph{softmax action selection}~\cite{SutBar18}, can easily be integrated into LCB sampling.

While our focus of LCB sampling is on exploiting ``good'' actions, we can still guarantee 
convergence towards an optimal strategy in the long run:

\begin{theorem}\label{thm:pac}
\Cref{alg:imdp-rl} with LCB sampling converges towards an optimal solution,
i.e., for $K\rightarrow\infty$ both $\underline{V}$ and $\overline{V}$ 
converge pointwise towards $V^{*}$, and their corresponding strategies
$\underline{\sigma}$ and $\overline{\sigma}$ converge towards optimal strategies.
\end{theorem}

Similar to how UCB sampling can provide PAC guarantees~\cite{AshKreWei19},
we can provide PAC guarantees for the value function bounds returned by \Cref{alg:imdp-rl} as
\Cref{thm:guarantees} guarantees that the solution 
is in the computed interval with high probability $1-\delta$
and \Cref{thm:pac} guarantees that the interval can become arbitrarily small
converging towards the optimal solution from both sides.

\subsection{Action scoping}

As another approach to compete with resource constraints, 
we propose to permanently remove unpromising actions from the 
learned IMDP model, forcing the agent to focus on a subset of enabled actions from the environment MDP.
We formalize this idea by setting the \emph{scope} of a state to the set of actions 
that the agent is allowed to perform in that state.

\subsubsection{Scope formation}
As depicted in \Cref{fig:workflow}, scopes are introduced after each episode
based on the samples of that episode.
Initially, all enabled actions are within a scope.
Removing an action $a$ from the scope in $s$ is enforced by modifying the interval transition function
$\hat{T}$ of $\cU$ to the zero interval function at $(s,a)$, i.e., $\hat{T}(s,a,s') = [0,0]$ 
for all $s'\in\Post(s,a)$. 
Scope formation has several notable advantages. First, removing action $a$ from a scope
in $s$ reduces the action space $\Act(s)$, leading to more sampling data for remaining actions
as $\sigma(s)(a)=0$ for all future episodes.
Further, the removal of actions may also reduce the state space in case states are only 
reachable through specific actions.
These positive effects of scoping come at its cost
of the algorithm not necessarily converging towards an optimal strategy anymore (cf. \Cref{thm:pac}).
The reason is in possibly removing an optimal action due to unfortunate sampling.

\subsubsection{Eager and conservative scopes}
We introduce two different scoping schemes: \emph{eager} 
and \emph{conservative}. Both schemes are tunable by a parameter $h\in\ ]0,1[$
that specifies the transition error tolerance
similar as $\eta = \delta/\Nt$ does in our IMDP construction (see \Cref{subsec:mb}).
Intuitively, while the formal analysis by means of \Cref{l:bounds} in \Cref{alg:imdp-rl}
guarantees $1{-}\delta$ correctness, we allow for different 
confidence intervals depending on $h$ when forming scopes. 
Here, greater $h$ corresponds to higher tolerance and hence smaller action scopes.

To define scopes, we introduce $\cU_h = (S,A,\imath,G,R,\hat{T}_h)$. That is, $\cU_h$
is an IMDP with the same topology as the internal model $\cU$, but allows an
error tolerance of $h$ in each transition.
We denote the corresponding solution to the interval Bellman equations of $\cU_h$ by
$\underline{V}_h$ and $\overline{V}_h$, respectively, and the quality functions as
$\underline{Q}_h$ and $\overline{Q}_h$.
Additionally, the \emph{mean quality function} $\dot{Q}$ is computed from the solution of the
Bellman equations on the maximum likelihood MDP $\dot{\cM} = (S,A,\imath,G,R,\dot{T})$ where 
$\dot{T}(s,a,s') = \#(s,a,s')/\#(s,a)$ are the maximum likelihood estimates of the transition probabilities.
Both functions can be computed, e.g., using standard value iteration approaches.

In state $s$ an action $a$ is \emph{eagerly} removed from its scope 
if $\dot{Q}(s,a) < \underline{V}_h(s)$, i.e., 
if the mean quality of $a$ is lower than the lower bound of the (presumably) best action.
The idea is that $a$ is most likely not worth exploring if its expected value 
is lower than what another action provides with high probability.
Likewise, an action $a$ is \emph{conservatively} removed from the scope of a state $s$ 
if $\overline{Q}_h(s,a) < \underline{V}_h(s)$, i.e.,
the upper bound quality of $a$ is lower than the lower bound of the (presumably) best action.
Here the idea is similar as for eager scoping but with a more cautious estimate on
the expected value from action $a$ (observe $\overline{Q}_h(s,a) > \dot{Q}(s,a)$).
Note that the parameter $h$ is only used as an error tolerance in $\cU_h$ in order
to reduce the action scopes. The bound $\underline{V}$ and $\overline{V}$ returned
in \Cref{alg:imdp-rl} still use an error tolerance of $\delta/\Nt$ per transition.


\section{Implementation and evaluation}\label{sec:evaluation}
To investigate properties of the algorithms presented, we developed a prototypical implementation in \python\ 
and conducted several experimental studies, driven by the following research questions:

\begin{enumerate}[label=\textbf{(RQ\arabic*)},leftmargin=*]
	\item\label{rq:sampling}
		How do UCB and LCB influence the quality of synthesized strategies?
	\item\label{rq:scoping}
		Does action scoping contribute to synthesize nearly optimal strategies 
		when limiting the number of samples?
\end{enumerate}

\subsection{Experiment setup}
We ran our experiments on various community benchmarks from the formal-methods and RL communities.
All our experiments were carried out using \python\ 3.9 on a MacBook Air M1
machine running macOS 11.5.2. For each system variant and scoping parameter, 
we learn $M$ strategies (i.e., run the algorithm $M$ times)
in $K{=}50$ episodes each with batch size $N$ as the number of state-action pairs 
that have a probabilistic successor distribution.
Plots show results averaged over the $M$ learned strategies.
We chose an error tolerance $\delta=0.1$, a total
of $k {\cdot} \lvert S\rvert$ value iteration steps in the $k$-th episode, and
an exploration of $\epsilon{=}0.1$.

\subsubsection{Models}
For an evaluation, we focus on two models: \racetrack\ and multi-armed bandits.
Results for all other experiments can be found in the appendix.

\begin{figure}[t]
 \begin{subfigure}[t]{1\textwidth}
  \centering
  \includegraphics[scale=1.50]{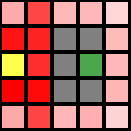}\qquad
  \includegraphics[scale=1]{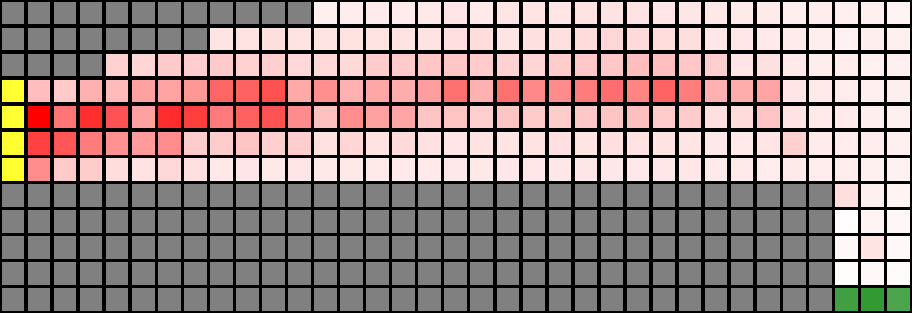}
   \caption{UCB sampling}
 \end{subfigure} \\[.5em]
 \begin{subfigure}[t]{1\textwidth}
  \centering
  \includegraphics[scale=1.50]{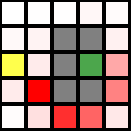}\qquad
  \includegraphics[scale=1]{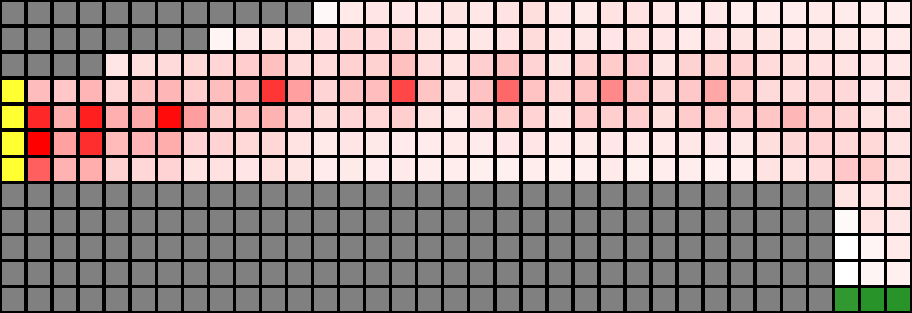}
   \caption{LCB sampling}
 \end{subfigure}
\caption{\racetrack\ exploration visualization of sampling methods\label{fig:racetrack_ucb_lcb}
 (tiles colored by start (yellow), goal (green), wall (dark gray), and visit frequency (red-white))}
\end{figure}

In \racetrack\ 
~\cite{BarBraSin95,PinZil14,Gro19,BaiDubHerKlaKluKoe20},
an agent controls a vehicle in a two-dimensional grid where 
the task is to reach a goal position from some start position, not colliding with wall tiles.
\Cref{fig:racetrack_ucb_lcb} depicts two example tracks from Barto et al.~\cite{BarBraSin95},
which we identify as ``small track'' (left) and ``big track'' (right).
At each step, the movement in the last step is repeated, possibly modified by 
$1$ tile in either direction, leading to $9$ possible actions in each state.
Environmental noise is modelled by changing the vehicle position by $1$ in each direction with small probability. 
We formulate \racetrack\ as RL problem by assigning goal states with one reward and all
other states with zero reward.
In the case that the vehicle has to cross wall tiles towards the new position, 
the run ends, not obtaining any reward.
In \racetrack\ experiments, we learn $M=10$ strategies constrained by $N=940$ sample runs.

The second main model is a variant of multi-armed bandits with one initial state having
$100$ actions, each with a biased coin toss uniformly ranging from $0.25$
to $0.75$ probability, gaining one reward and returning to the initial state. 
Here, we learn $M=100$ strategies constrained by $N=101$ sample runs.

\subsection{Sampling methods \ref{rq:sampling}}
We investigate the differences of UCB and LCB sampling 
within \racetrack.

\subsubsection{State-space coverage}
UCB and LCB sampling differ notably in covering the state space while learning.
With UCB sampling, actions with high uncertainty are more likely to be executed,
lowering their upper bound and thus increasing the chance of
other actions with higher uncertainty in the next sample run.
Hence, UCB sampling leads to exploration of many actions and thus to a high 
coverage of the state space.
In contrast, LCB sampling increases confidence in one particular action
shown viable in past samples, leading to sample the same action sequences more often.
Hence, LCB sampling is likely to cover only those states visited by
one successful sampling strategy, showing low coverage of the state space.
This can be also observed in our experiments.
\Cref{fig:racetrack_ucb_lcb} shows the frequency of visiting positions in the
small and big example tracks, ranging from high (red) to low (white) frequencies.
Both tracks already illustrate that UCB sampling
provides higher state-space coverage than LCB sampling.
The small track is symmetric and for each
strategy striving towards a lower path, there is a corresponding equally performing strategy
towards an upper path. UCB sampling treats both directions equally, while
the LCB sampling method in essential learns one successful path and increases its confidence, 
which is further reinforced in the following samples. 
reached by one of the symmetric strategies.
\begin{figure}[t]
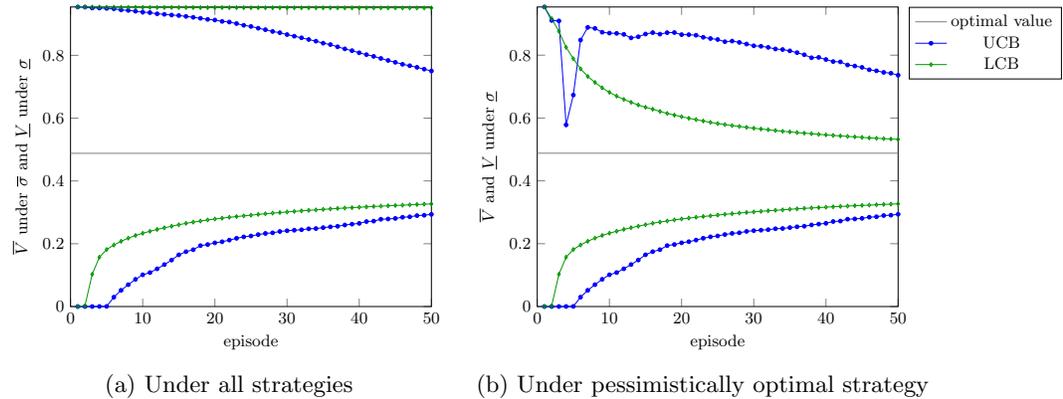

 \begin{subfigure}[t]{.5\textwidth}
  \centering\scalebox{0.7}{\input{img/LCBvsUCB}}
   \caption{Under all strategies}
 \end{subfigure}
 \,
 \begin{subfigure}[t]{.5\textwidth}
  \centering\scalebox{0.7}{\input{img/LCBvsUCB2}}
   \caption{Under pessimistically optimal strategy}
 \end{subfigure}
\caption{Comparison of obtained bounds for strategies\label{fig:tiny_ucb_lcb_graphs}}
\end{figure}

\subsubsection{Robustness}
A further difference of the sampling methods is in dealing with less-explored situations,
where UCB sampling is likely to explore new situations but LCB sampling prefers
actions that increase the likelihood of returning to known states of an already 
learned viable strategy. This is due to those states having
smaller confidence intervals and thus a greater lower bound
on the value and quality functions.
\Cref{fig:racetrack_ucb_lcb} shows this effect in the frequency plot of the big track:
LCB sampling leads to only few isolated positions with high visit frequencies, while
UCB shows a trajectory of visited positions.

\subsubsection{Guaranteed bounds}
The different characteristics of UCB and LCB sampling can also be observed
during the learning process in the small track. 
In \Cref{fig:tiny_ucb_lcb_graphs} on the left we show 
$\underline{V}$ and $\overline{V}$ after each episode. 
Note that these bounds apply to different 
strategies, i.e., the optimistically optimal strategy $\underline{\sigma}$ maximizes $\underline{V}$,
while the pessimistically optimal strategy $\overline{\sigma}$ maximizes $\overline{V}$. 
Here, LCB provides values $\underline{V}$ closer to the optimum and, 
due to its exploitation strategy, gains more confidence in its learned strategy.
However, unlike UCB sampling, it cannot improve on $\overline{V}$ significantly, 
since parts of the environment remain mostly unexplored.
We plot bounds under the single fixed strategy $\underline{\sigma}$ on the right. 
After 50 episodes, UCB then can provide value function bounds $[0.29,0.75]$,
while LCB provides $[0.33,0.53]$, being more close to the optimal
value of $0.49$.

LCB is also favourable under limited sampling access, e.g., in
(mostly) symmetric environments as the small track: UCB explores the symmetry and 
requires at least double samples for achieving a similar confidence on the learned strategy.

\hframe{
Concerning \ref{rq:sampling}, we showed that LCB sampling can provide better strategies with
high confidence than UCB sampling, while UCB sampling shows better bounds 
when ranging over all strategies.
}

\subsection{Impact of scoping \ref{rq:scoping}}
We now investigate the impact of action scoping and its parameter $h$
on the multi-armed bandit experiment. 
    
    \begin{figure}[t]
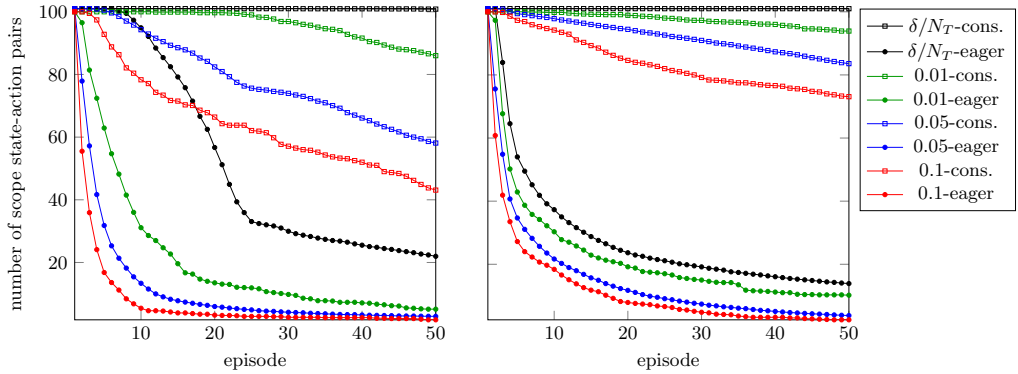

 \begin{subfigure}[t]{.5\textwidth}
  \centering\scalebox{.8}{\input{img/UCB_num_habits}}
 \end{subfigure} \,
 \begin{subfigure}[t]{.5\textwidth}
  \centering\scalebox{.8}{\input{img/LCB_num_habits}}
 \end{subfigure}
\caption{Action-space reduction by action scoping (UCB left, LCB right)\label{fig:num_habits_ucb_lcb}}
\end{figure}

\subsubsection{Action-space reduction}
\Cref{fig:num_habits_ucb_lcb} shows the number of state-action pairs in the IMDP
after each episode w.r.t. UCB and LCB sampling. Here, eager and conservative action 
scoping is considered with various scoping parameters $h$.
As expected, more actions are removed for greater $h$.
Since $\dot{Q}(s,a) \leqslant \overline{Q}_h(s,a)$, eager scoping leads to
more actions being removed than conservative scoping (cf. eager plots in the lower part 
of the figures). Observe that the choice of eager or 
conservative scoping has more impact than the choice of $h$.
In terms of the sampling method we observe that for conservative scoping with UCB sampling more actions 
are removed from scopes than with LCB sampling. A possible explanation is that 
in LCB sampling, suboptimal actions do not acquire enough samples to 
sufficiently reduce the upper bound of their expected reward.

\begin{figure}[t]
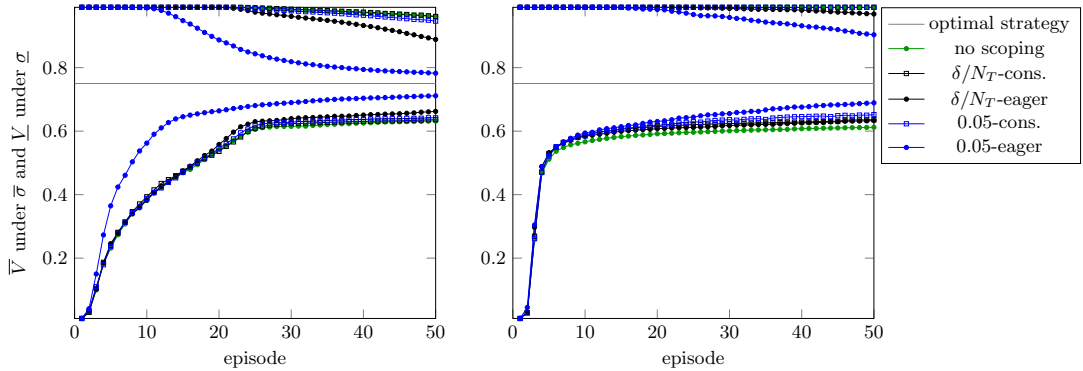

 \begin{subfigure}[t]{.5\textwidth}
  \centering\scalebox{.8}{\input{img/appendix/Bandit25-75UCB_bounds}}
 \end{subfigure} \,
 \begin{subfigure}[t]{.5\textwidth}
  \centering\scalebox{.8}{\input{img/appendix/Bandit25-75LCB_bounds}}
 \end{subfigure}
\caption{Bounds of the subsystem obtained by scoping (UCB left, LCB right)\label{fig:habit_bounds_Bandit25-75}}
\end{figure}

\begin{figure}[t]
 \begin{subfigure}[t]{.5\textwidth}
  \centering\scalebox{.8}{\input{img/appendix/Bandit25-75UCB_corr_bounds}}
 \end{subfigure} \,
 \begin{subfigure}[t]{.5\textwidth}
  \centering\scalebox{.8}{\input{img/appendix/Bandit25-75LCB_corr_bounds}}
 \end{subfigure}
\caption{Bounds for pessimistically optimal strategy (UCB left, LCB right)\label{fig:habit_corr_bounds_Bandit25-75}}
\end{figure}

\subsubsection{Strategy bounds}
Next, we investigate the bounds obtained by the strategies returned by \Cref{alg:imdp-rl}. 
For brevity, we focus here on the cases $h{=}\delta/\Nt$ and $h{=}0.05$. 
Our results are plotted in \Cref{fig:habit_bounds_Bandit25-75} and \Cref{fig:habit_corr_bounds_Bandit25-75}
for $\underline{V}$ and $\overline{V}$ on the subsystem obtained by applying action scopes with 
both $\underline{\sigma}$ and $\overline{\sigma}$ and solely $\underline{\sigma}$, respectively.
For UCB sampling, we observe that bounds tighten faster the 
more actions are removed from scopes and reduce the system size,
i.e., particularly for eager scoping and for $h{=}0.05$.
For LCB, scopes do not have such a drastic influence, since actions are only leaving the scope 
if there is an alternative action with high $\underline{V}$, in which case the latter 
action is sampled mostly anyway.

\begin{figure}[t]
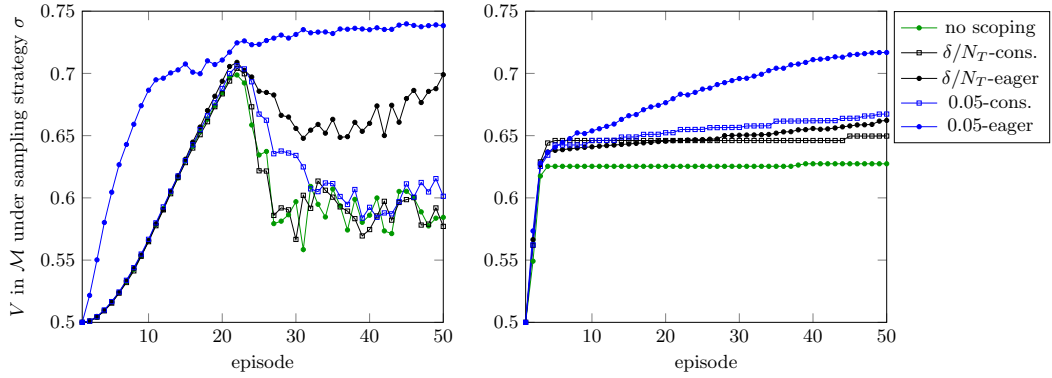

 \begin{subfigure}[t]{.5\textwidth}
  \centering\scalebox{.8}{\input{img/appendix/Bandit25-75UCB_real}}
 \end{subfigure} \,
 \begin{subfigure}[t]{.5\textwidth}
  \centering\scalebox{.8}{\input{img/appendix/Bandit25-75LCB_real}}
 \end{subfigure}
\caption{Expected total reward w.r.t. sampling strategy $\sigma$ (UCB left, LCB right)\label{fig:real_strategy_Bandit25-75}}
\end{figure}
\subsubsection{Sampling strategy quality}
In \Cref{fig:real_strategy_Bandit25-75} we plot the expected total 
reward of the employed sampling strategy $\sigma$ in \Cref{alg:imdp-rl} after each episode. 
Eager scoping tremendously improves the quality of the sampling strategy for both
UCB and LCB sampling.
For the UCB strategy we observe an initial monotonic increase of the online performance
that that eventually drops off. This is because a lot of actions cannot
improve on the trivial upper bound of 1 until a lot of samples are acquired.
In the first roughly 20 episodes the number of actions with trivial upper bound decreases
as more samples are collected. Additionally, the actions with upper bound less than 1
are those which were less successful in the past, i.e., likely suboptimal actions, explaining the monotonic increase.
Afterwards, the fluctuations within UCB sampling stem from the fact that after reducing all upper bounds below 1,
UCB sampling may result in a fully deterministic sampling strategy that only updates after each episode.
Hence, exploring a particular action takes the full $N$ runs of an episode, even if the action is suboptimal
and only has a high upper bound due to a lack of samples.
Especially here, scoping helps to eliminate such actions and avoids sampling them for
a full episode just to confirm the action was indeed suboptimal.
For LCB sampling the better performance with scoping is due to suboptimal
actions being removed and thus not eligible in the exploration step with probability $\epsilon$.

\subsubsection{Subsystem bounds}
With the introduction of scopes, our RL algorithm is not guaranteed to converge to 
optimal values. To determine whether optimal actions are removed from scopes in practice, 
we compute the optimal strategy within the subsystem 
generated by only considering actions within the computed scope. 
Note that for the transition function we are using the exact probabilities 
as in the environment MDP. The results are given in \Cref{tbl:subsystem_bounds_bandit}. 
Without scope reduction the subsystem is just the entire environment. 
When introducing scopes, we did not remove the optimal action via conservative scoping a single time
with either sampling method, even for $h=0.05$.
Only with eager scoping we saw the optimal action being removed from the scope, but the optimal strategy
in the scoped subsystem still performs reasonably well compared to the overall optimal strategy. 
The fact we observe this only with eager scoping is
not surprising, as removing
more actions from scopes (recall \Cref{fig:num_habits_ucb_lcb})
of course increases the chance of removing the optimal action in a state.

\begin{table}[t]
\caption{Values of the optimal strategy in the multi-armed bandit model\label{tbl:subsystem_bounds_bandit}}
\resizebox{\columnwidth}{!}{%
\begin{tabular}{l|c|clcl||clcl}
\toprule
                                                            & no             & \multicolumn{2}{c|}{UCB-cons.}                                 & \multicolumn{2}{c||}{UCB-eager}                                    & \multicolumn{2}{c|}{LCB-cons.}                                 & \multicolumn{2}{c}{LCB-eager}                                    \\
                                                            & \multicolumn{1}{l|}{scoping} & \multicolumn{1}{l|}{$h{=}\delta/\Nt$} & \multicolumn{1}{l|}{$h{=}0.05$} & \multicolumn{1}{l|}{$h{=}\delta/\Nt$} & $h{=}0.05$                  & \multicolumn{1}{l|}{$h{=}\delta/\Nt$} & \multicolumn{1}{l|}{$h{=}0.05$} & \multicolumn{1}{l|}{$h{=}\delta/\Nt$} & $h{=}0.05$                 \\ \midrule
value & 0.75                  & \multicolumn{1}{c|}{0.75}         & \multicolumn{1}{c|}{0.75}     & \multicolumn{1}{c|}{0.75}         & \multicolumn{1}{c||}{0.742} & \multicolumn{1}{c|}{0.75}         & \multicolumn{1}{c|}{0.75}     & \multicolumn{1}{c|}{0.746}         & \multicolumn{1}{c}{0.739}
\\ \bottomrule
\end{tabular}
}
\end{table}
\hframe{
For \ref{rq:scoping}, we conclude that for both UCB and LCB sampling, scoping and
especially eager scoping significantly improves the quality of learned strategies
after few samples, while only slightly deviating from the optimal strategy.
In the UCB setting, scoping leads to further exploitation and thus better bounds.
}

\subsection{Further examples}
We ran our algorithms on several other environment MDPs from the RL and formal-methods communities. 
The bounds of the expected reward obtained for the reduced subsystem, as well as for the strategy
maximizing the lower bound, are given in the appendix (cf. \Cref{tbl:experiments}).
In general, the strategy learned from LCB yields equal or higher lower bounds and tighter bounds for single 
strategies, while UCB sampling gives tighter bounds for the entire system. Employing action scoping generally
tightens the bounds further with the eager scoping emphasizing this effect.
The margins of the differences vary between the examples. In general, both 
LCB sampling and scoping have biggest impact on large action spaces
and on models with high probability deviations 
such as with small error probabilities.
On the flipside, we observed that LCB performs poorly when mostly or fully deterministic actions or even runs
are available, as those incur little uncertainty and thus tend to have relatively large lower bounds even with few samples.


\section{Concluding remarks}\label{sec:conclusions}

We devised novel model-based RL algorithms that are inspired by efficient human reasoning
under time constraints.
Similar to humans tending to stick and return to known situations during strategy learning,
LCB exploration favors to return to states with high confidence and proceeding with
viable learned strategies.
On the action level, scoping implements a reduction of the exploration space
as humans do when favoring known actions without further exploration.
As for humans acting under resource constraints, both ingredients have been shown to yield
better strategies after few sample runs than classical RL methods, especially
when choosing high scoping parameters that foster action scoping.
While our methods synthesize good strategies faster, an optimal strategy
might be not achievable in the limit.
We would like to emphasize that also using IMDPs as internal model for RL is a result by
itself, incorporating the knowledge about action scopes and confidences in the learned
probabilities of the environmental MDP model.

We mainly discussed applications of our techniques in the setting of reinforcement learning.
Nevertheless, they can well be utilized also in the formal methods domain, 
providing an existential variant for statistical model checking of MDPs,
asking for the existence of a strategy to reach a goal with accumulating a certain reward.

In future work, we will further investigate adaptations of our algorithms to reflect
main properties of human reasoning, e.g., by establishing real-world neuroscientific 
experiments that involve humans. Our vision is to use formal methods
to support explanations of human behaviors, to which we see the 
present paper providing a first step in this direction~\cite{BaiDubFun21}.
Future adaptations of our proposed algorithm may focus on more sophisticated
exploration schemes than dithering, such as softmax selection or exploration bonuses,
that allow for deeper exploration of the environment \cite{SutBar18}.
While not the main focus of this paper, it is well possible to extend our approach also 
to a black-box setting, i.e., without knowledge about the topology of the MDP, using 
similar techniques as in \cite{AshKreWei19}. 
One advantage of using the gray-box setting is in also ensuring applicability to the
instance of infinite state MDPs with finitely many actions if this MDP can be effectively
explored. For this, it suffices to consider only a finite fragment of the MDP given
that our algorithms restrict the sample lengths to a fixed bound.

\clearpage
\bibliographystyle{splncs04}
\bibliography{main}

\appendix
\newpage

\section{Appendix}
In this appendix, we provide content that had to be omitted from the main paper
due to the lack of space.

\subsection{Environment approximation}
While the approximation approach by Ashok et al. \cite{AshKreWei19} does 
not explicitly operate on IMDPs, it can be expressed in our setting by
\[
	\hat{T}(s,a,s')\quad =\quad \left[\max\big(p_{min},\frac{\#(s,a,s')}{\#(s,a)}-c'\big),1\right]
\]
where $c'=\sqrt{\frac{\log\delta_T}{-2\#(s,a)}}$.
The latter is the confidence interval obtained
by applying the Hoeffding inequality \cite{Hoeffding} that
guarantees $(1{-}\delta_T)$-correctness for each transition triple
$(s,a,s') \in S {\times} A {\times} S$ when viewing one sample of action 
$a$ in state $s$ as a Bernoulli trial.
Their method takes only the lower bound of the transition probability
into account, which we extend to also consider the upper bound.

We now prove the correctness of our definition of 

\begin{lemma}\label{lemma:approximation}
Let $\cM=(S,A,\imath,G,R,T)$ be an MDP,  and $\eta \in ]0,1[$.
Assume that for a fixed state $s\in S$ an action $a\in \Act(s)$ has been sampled $\#(s,a) \in \Nat_{>0}$ times and
successor $s'$ was observed $\#(s,a,s')$ times. Then, with probability $1-\eta$ it holds that
$$ T(s,a,s') \in \left[\frac{\#(s,a,s')}{\#(s,a)}-c,\frac{\#(s,a,s')}{\#(s,a)}+c\right]$$
where $c=\sqrt{\frac{\log\eta/2}{-2\#(s,a)}}$
\end{lemma}

\paragraph{Proof.}

We can directly bound the probability of $T(s,a,s')$ being outside the interval for a given $s,a,s'$
by applying the two-sided Hoeffding bound, proving the Lemma.

\begin{align*}
P \left( \left| T(s,a,s') - \frac{\#(s,a,s')}{\#(s,a)} \right|  \geqslant c \right) & \leqslant 2 \exp \left( -2\#(s,a) c^2 \right) \\
& = 2 \exp \left( -2\#(s,a) \sqrt{\frac{\log\eta/2}{-2\#(s,a)}}^2 \right) \\
& = \eta
\end{align*}
\qed

Using this result we can also easily show that $\cM\in\cU=(S,A,\imath,G,R,T_\eta)$
where $\eta = \delta / \Nt$. As $\cM$ and $\cU$ by definition have the same graph structure
and rewards, we only need to check that $T(s,a,s')\in T_\eta(s,a,s')$ for all $s\in S, a\in\Act(s)$ and
$s'\in\Post(s,a)$.

First, note that for a non-probabilistic transition (i.e., $s,a$ with $\Post(s,a)=\{s'\}$)
the statement also holds trivially as $T(s,a,s')=\frac{\#(s,a,s')}{\#(s,a)}\in T_\eta(s,a,s')$.

Using \Cref{lemma:approximation} the probability that $T(s,a,s')$ is outside the interval 
for any of the $\Nt$ probabilistic transitions can thus be bounded by $\eta \Nt = \delta$. 

\subsection{Building minimizing and maximizing instantiations}

\begin{algorithm}[t]
\SetAlgoLined
	\DontPrintSemicolon
	\SetKwInOut{Input}{Input}\SetKwInOut{Output}{Output}
\Input{IMDP $\cU=(S,A,\imath,G,R,\hat{T})$, value function estimate $\hat{V}$}
\Output{value minimizing transition function $T^{-}$}
\BlankLine
\ForAll{$s\in S$}{
	\ForAll{$a\in\Act(s)$}{
		\lForAll{$s'\in\Post(s,a)$}{$T^{-}(s,a,s') := \underline{T}(s,a,s')$}
		\While{$\sum_{s'\in\Post(s,a)} T^{-}(s,a,s') < 1$}{
			\If{$1-\sum_{u\in\Post(s,a)} T^{-}(s,a,u) > \overline{T}(s,a,s')-\underline{T}(s,a,s')$}{
				$T^{-}(s,a,s') := \underline{T}(s,a,s')$
			}
			\Else{
				$T^{-}(s,a,s') := 1-\sum_{u\in\Post(s,a),u\neq s'} T^{-}(s,a,u) $
			}
		}
	}
}
\Return{$T^{-}$}
\caption{MINIMIZING\_TRANSITIONS($\cU, \hat{V}$) \label{alg:minimizing_transitions}}
\end{algorithm}

A core idea in the adapted value iteration algorithm for IMDPs is to find an instantiation
that minimizes or maximizes the value in each state with respect to a previously computed
value function \cite{Givan00,Wu08}.

Formally, we can state the problem as follows: Given an IMDP $\cU=(S,A,\imath,G,R,\hat{T})$ with 
value function estimate $\hat{V} : S \rightarrow \Intv(S)$, choose for each state $s$
and action $a\in \Act(s)$ a transition instantiation $t_s^a\in T_s^a$ such that
$\hat{Q}(s,a)=\sum_{s'\in\Post(s,a)}\hat{V}(s') t_s^a$ is minimized (resp. maximized).

As the instantiation of $t_s^a$ only impacts any $\hat{Q}(s,a)$, and $\hat{Q}$ for no other
parameters,  this can be optimized locally in a straight forward way.  We exemplify this for
the minimization case in \Cref{alg:minimizing_transitions}. 
The maximization case is analogous. First, in an effort to ensure that 
$t_s^a\in T_s^a$, we shift the minimum amount of probability mass into each successor state,
i.e., we initialize the value of $t_s^a(s')$ to $\underline{T}(s,a,s')$ for each $s'\in\Post(s,a)$.
We then distribute the remaining probability mass into the state $s^{*}\in\Post(s,a)$ for which
$\hat{V}(s^{*})$ is minimal (resp. maximal) until either the total probability mass of $t_s^a$
reaches $1$, or until $t_s^a(s^{*})$ reaches $\overline{T}(s,a,s')$.
If the former is the case, we are done. Otherwise, we repeat the procedure with the state
out of the other states for which $\hat{V}$ is minimal (resp. maximal).

As we initialized all transition probabilities with $\underline{T}(s,a,s')$ and only increase them,
but to no more than $\overline{T}(s,a,s')$, and as we stop when $t_s^a$ reaches a probability
mass of $1$ we guarantee that $t_s^a\in T_s^a$.  The result also clearly minimizes $\hat{Q}(s,a)$.

\subsection{Value iteration for IMDPs}

\begin{algorithm}[t]
\SetAlgoLined
	\DontPrintSemicolon
	\SetKwInOut{Input}{Input}\SetKwInOut{Output}{Output}
\Input{IMDP $\cU=(S,A,\imath,G,R,\hat{T})$, parameter $k$}
\Output{Lower and upper bound value functions $\underline{V}$ and $\overline{V}$}
\BlankLine
$\underline{V}, \overline{V} = \text{INITIALIZE\_BOUNDS}(\cU)$ \;\label{line:init}
\RepTimes{$k\cdot\lvert S \rvert$}{
	$T^{-}:=$ MINIMIZING\_TRANSITIONS$(\cU,\underline{V})$\;
	$T^{+}:=$ MAXIMIZING\_TRANSITIONS$(\cU,\overline{V})$\;
	\ForAll{$s\in S$}{
		$\underline{V}'(s) := R(s) + \max_{a\in \Act(s)} \sum_{s' \in S} \underline{V}(s')\cdot T^{-}(s,a,s')$\;\label{line:vu}
		$\overline{V}'(s) := R(s) + \max_{a\in \Act(s)} \sum_{s' \in S} \overline{V}(s')\cdot T^{+}(s,a,s')$\;
	}
	$\underline{V} := \underline{V}'$\;
	$\overline{V} := \underline{V}'$\;
}
\Return{$\underline{V}',\overline{V}'$}
\caption{COMPUTE\_BOUNDS($\cU, k$) \label{alg:uncertain_vi}}
\end{algorithm}

We briefly recall the IMDP value-iteration algorithm outlined in \Cref{alg:uncertain_vi}
first developed by Givan et al.  \cite{Givan00} for discounted rewards, and later applied by
Wu et al. \cite{Wu08} for positive rewards.\\
An important observation for that is that for any IMDP $\cU$ there is an instantiation that is 
minimizing (or maximizing) the value function for all states simultaneously.
The main idea behind the algorithm is then to guess a total order $\leq_V$ over all states $s \in S$
and assume that $s \leq_V s' \iff V(s) \leqslant V(s')$. Given this assumption one can easily perform a
value iteration step by first computing the minimizing transition instantiation for a state $s$ and an action $a$:
Transitions are initially instantiated with the lower bound of $\hat{T}(s,a,s')$ for each
successor state $s'\in \Post(s,a)$ and then probability mass is added to the transitions to the successors 
which are minimal w.r.t. $\leq_V$ until the sum of the instantiated probabilities reaches $1$. 
The computation for the maximizing MDP is analogous.
Following that, the order $\leq_V$ is redefined according the value function computed in the VI step. 

In total, we perform $k\cdot\lvert S \rvert$ such value iteration steps where $k$ gives the number of episodes that
have passed. \footnote{Technically any function $f\colon \Nat \rightarrow \Nat$ with the restriction that
$\lim_{k\rightarrow\infty} f(k)=\infty$ could be used here, but we choose $k\cdot\lvert S \rvert$ as linear growth
in $k$ avoids unnecessarily many iterations and scaling by $\lvert S \rvert$ adapts the algorithm to the 
environment size}

Recall that there is no stopping criterion with formal guarantees.
Towards proving that the algorithm can be seen as an any-time algorithm, we first show the following Lemma:

\begin{lemma}\label{lemma:vi}
Let $\underline{V}^{*}$ and $\overline{V}^{*}$ be the solution of the 
interval Bellman equations of an IMDP $\cU$, respectively, and
$([\underline{V}_i(s),\overline{V}_i(s)])_{t\in\Nat}$ be a sequence of interval value functions s.t.
$[\underline{V}_0(s),\overline{V}_0(s)] \supseteq [\underline{V}^{*}(s),\overline{V}^{*}(s)]$ for all $s\in S$ 
and $\dot{V}_{i+1}(s)$ where $\dot{V}\in\{ \underline{V},\overline{V} \}$ is computed
according to \Cref{alg:uncertain_vi}, i.e.,
\begin{align*}
\underline{V}_{i+1}(s) & = R(s) + \max_{a\in \Act(s)} \sum_{s' \in S} \underline{V}_i(s')\cdot T^{-}(s,a,s') \\
\overline{V}_{i+1}(s) & = R(s) + \max_{a\in \Act(s)} \sum_{s' \in S} \overline{V}_i(s')\cdot T^{+}(s,a,s')
\end{align*} 
Then $[\underline{V}_i(s),\overline{V}_i(s)] \supseteq [\underline{V}^{*}(s),\overline{V}^{*}(s)] $ for all $i\in\Nat, s\in S$.
\end{lemma}

\paragraph{Proof.}
We first show the statement for the lower bound, i.e., that $\underline{V}_i(s) \leqslant \underline{V}^{*}(s)$.

By induction.  The base case $\underline{V}_0$ holds by assumption.

For the induction step, we assume $\underline{V}_i$ satisfies the condition, i.e.,
$\underline{V}_i(s) \leqslant \underline{V}^{*}(s)$ for all $s\in S$.

First, notice that by definition of MINIMIZING\_TRANSITIONS, we have
$T^{-}(s,a) = \argmin_{t_s^a \in T_s^a} \sum_{s' \in S} \underline{V}(s')\cdot t_s^a(s')$.
This is the case since the function shifts as much probability mass as possible
into states with a low value function in the previous step. 

Then,by induction hypothesis and because $t^a_s(s')\geqslant 0$ the following chain holds:

\begin{align*}
\underline{V}^{*}(s) & = \max_{a\in \Act(s)} \min_{t_s^a \in T_s^a} \sum_{s' \in S} \underline{V}^{*}(s')\cdot t_s^a(s') \\
& \geqslant  \max_{a\in \Act(s)} \min_{t_s^a \in T_s^a} \sum_{s' \in S} \underline{V}_i(s')\cdot t_s^a(s') \\
& = \underline{V}_{i+1}(s)
\end{align*}

The analogous case for upper bounds can be proven in the same way, i.e.,
for any $i\in \mathbb{N}$ and $s\in S$
we can guarantee that $\overline{V}_i(s)\geq\overline{V}^{*}(s)$.

From the inequalities for both bounds we can deduce that indeed 
$[\underline{V}_i(s),\overline{V}_i(s)] \supseteq [\underline{V}^{*}(s),\overline{V}^{*}(s)]$
for all $t\in \mathbb{N}$.\qed

\paragraph{}\noindent
Note that this requires a safe upper and lower bound,  $\underline{V}_0(s)$ and $\overline{V}_0(s)$, 
of the value function in each state $s \in S$.
For reachability tasks the trivial bounds $0$ and $1$ suffices while 
for general reward structures the topology of the environment along with $p_{min}$ can 
be used to compute such bounds \cite{BeiKleLeu17}.

The sequences $(\underline{V}_i)_{i\in \Nat}$ and $(\overline{V}_i)_{i\in \Nat}$ are not necessarily monotonic.
However, one could adapt the definition to 
$$ \underline{V}_{i+1}(s) = \max \{ \underline{V}_i(s), \underline{W}_i \}$$
with
$$\underline{W}_i(s) = R(s) + \max_{a\in \Act(s)} \sum_{s' \in S} \underline{V}_i(s')\cdot T^{-}(s,a,s') $$
and analogously (using the minimum between $\overline{W}$ and $\overline{V}$) for the upper bound, to enforce monotonicity. 
Under this definition \Cref{lemma:vi} still holds, as the proof already handles the case
$\underline{W}_i(s) \geqslant \underline{V}_i(s)$, and the case $\underline{W}_i(s) < \underline{V}_i(s)$
can be proven directly from the induction hypothesis as it implies $\underline{V}_{i+1}(s) = \underline{V}_i(s)$

\begingroup
\def\thetheorem{\ref{thm:guarantees}}
\begin{theorem}
Let $V^{*}$ be the solution to the Bellman equations of a given MDP $\cM$. 
Then for all $\delta\in\ ]0,1[$ and $K,N\in\Nat$
the value function bounds $\underline{V}$ and $\overline{V}$ returned by 
\textup{IMDP\_RL($\cM,\delta,K,N$)} as of \Cref{alg:imdp-rl} contain $V^{*}$ 
with probability at least $1-\delta$, 
i.e., $\Prob\left(\underline{V}(s) \leqslant V^{*}(s) \leqslant \overline{V}(s)\right)\geqslant 1-\delta$
for all $s\in S$. 
\end{theorem}
\addtocounter{theorem}{-1}
\endgroup

\paragraph{Proof.}
By \Cref{lemma:approximation} we have $\cM\in [\cU]$ with probability at least $1-\delta$.
Assume $\cM\in[\cU]$ and let $\underline{V}^{*}$ and $\overline{V}^{*}$ be the solutions to
the interval Bellman equations. Then, clearly $\underline{V}^{*}(s) \leqslant V^{*}(s) \leqslant \overline{V}^{*}(s)$
for all $s\in S$ and further by \Cref{lemma:vi} $\underline{V}_i(s) \leqslant V^{*}(s) \leqslant \overline{V}_i(s)$
for all $t\in\Nat$ where $\underline{V}_i(s)$ and $\overline{V}_i(s)$ denote the computed 
lower and upper bounds of value function in the $t$-th iteration step.  Note that \Cref{lemma:vi}
is applicable since we pick safe initial bounds in \Cref{line:init} in \Cref{alg:uncertain_vi}.

Hence, $\Prob\left(\underline{V}(s) \leqslant V^{*}(s) \leqslant \overline{V}(s)\right)\geqslant 1-\delta$
for all $s\in S$. \qed

\subsection{LCB sampling}

\begin{algorithm}[t]
	\SetAlgoLined
	\DontPrintSemicolon
	\SetKwInOut{Input}{Input}\SetKwInOut{Output}{Output}
	\Input{An IMDP $\cU=(S,A,\imath,G,R,\hat{T})$ with lower bounds on its value function $\underline{V}$, exploration parameter $\epsilon$}
	\Output{a strategy $\sigma$}
	\BlankLine
	
	$\leq_V := \{ (s,s') \mid V(s) \leqslant V(s') \}$\;
	\ForAll{$s\in S$}{
	\ForAll{$a\in \Act(s)$}{
		$T^{-} :=$ MINIMIZING\_TRANSITIONS$(\cU,\leq_V)$\;
		$\underline{Q}(s,a) := R(s) + \sum_{s'\in S} \underline{V}(s)\cdot T^{-}(s,a,s')$\;
		$\sigma(s,a) := \epsilon / \lvert \Act(s) \rvert$
	}
	$a^{*} := \argmax_{a\in \Act(s)} Q(s,a)$\;
	$\sigma(s,a^{*}) :=  \sigma(s,a^{*}) + (1-\epsilon)$
	}

	\Return $\sigma$
	\caption{UPDATE\_LCB\_STRATEGY($\cU, \underline{V},\epsilon)$\label{alg:policy}}
\end{algorithm}

After computing the bounds of the value function for a given dataset over the unknown MDP (cf. \Cref{alg:uncertain_vi}),
we can utilize the obtained bounds to define the strategy to be used to sample the MDP in further episodes.
We outline this procedure in \Cref{alg:policy}.  Similar to the idea behind value iteration on IMDPs,
we first define a state ordering according to the given lower bounds of the value function.
We then also compute the instantiation that minimizes the value function in all states (cf. \cref{alg:minimizing_transitions}).
This yields the lower bound of the expected reward for each state-action pair.
While each action $a\in\Act(s)$ is allocated a probability of $\epsilon/\Act(s)$, 
the action $a^{*}$ with the highest lower bound is allocated the the remaining $(1-\epsilon)$ probability mass
of the strategy is state $s$.

As described in the main part of the paper, the $\epsilon$-sampling serves as an exploration mechanism
while at the same time guaranteeing the algorithm converges in the long run.

\begingroup
\def\thetheorem{\ref{thm:pac}}
\begin{theorem}
\Cref{alg:imdp-rl} with LCB sampling converges towards an optimal solution,
i.e., for $K\rightarrow\infty$ both $\underline{V}$ and $\overline{V}$ 
converge pointwise towards $V^{*}$, and their corresponding strategies
$\underline{\sigma}$ and $\overline{\sigma}$ converge towards optimal strategies.
\end{theorem}
\addtocounter{theorem}{-1}
\endgroup

\paragraph{Proof.}
Let $\cM$ be an MDP and $\cU$ the IMDP computed according to \Cref{alg:update}, calling \Cref{alg:uncertain_vi}.  
We first show the claim that as $K\rightarrow\infty$, the computed bound of the value function
$\underline{V}$ and $\overline{V}$ converge towards the correct solution to the interval Bellman equations, i.e.,
for all $s\in S$:

\begin{align*}
\underline{V}(s) & = \min_{\cM\in [\cU]} V_\cM(s) \\
\overline{V}(s) & = \max_{\cM\in [\cU]} V_\cM(s)
\end{align*}

As that the main loop, i.e., the value iteration step, is executed $k\cdot \lvert S \rvert$ times in the $k$-th episode,
and the value iteration step is a point-wise contraction mapping
(cf.  \cite[Lemma 5]{Wu08}), we can apply Banach fixed-point theorem to 
show that the value iteration converges as $k\rightarrow\infty$. 

Further, it is known that there are globally reward-minimizing (resp. reward maximizing) MDP $\cM^{-}$ and $\cM^{+}$, i.e.,

\begin{align*}
\underline{V}(s) & = V_{\cM^{-}}(s) \\
\overline{V}(s) & = V_{\cM^{+}}(s)
\end{align*}

for all $s\in S$ \cite{Givan00,Wu08}. As both $\cM^{-}$ and $\cM^{+}$ have the same graph structure as $\cM$,
they are contracting as well and thus there is a unique fixed point of the Bellman equations \cite{BerTsi91}.
Hence, the value iteration converges the the (unique) solution to the Bellman equations.

Next, we show that $\underline{V}$ and $\overline{V}$ converge towards $V^{*}$ under LCB sampling.

It is well known that, as long as each run has the chance to visit any state,
an $\epsilon$-greedy sampling strategy is guaranteed to visit each state infinitely 
often. More precisely, the probability of reaching any state $s$ in $L \geqslant \lvert S \rvert$ steps
can be bounded from below by a constant $(p_{min} \frac{\epsilon}{\lvert A \rvert})^{\lvert S \rvert}>0$. 
Thus, the expected number of times $s$ is visited as well as the number of times any action
$a\in\Act(s)$ is sampled in $S$ go towards infinity.
Hence, the confidence interval size approaches 0, i.e., $\lim_{t\rightarrow\infty}c(s,a) = 0$. 
Further, by the law of large numbers, 
$\lim_{t\rightarrow\infty}\frac{\#(s,a,s')}{\#(s,a)} = T(s,a,s')$ and thus
$\lim_{t\rightarrow\infty}\hat{T}(s,a,s') = [T(s,a,s'), T(s,a,s')]$.

In particular, this implies that
\[
	\lim_{t\rightarrow\infty}\sum_{s\in S}\underline{T}(s,a,s')\quad = \quad
\lim_{t\rightarrow\infty}\sum_{s\in S}\overline{T}(s,a,s') \quad =\quad 1
\] and thus the
distributions $T^{-}$ and $T^{+}$ in COMPUTE\_BOUNDS($\cU)$,
converge pointwise towards the true distribution $T$ which means the solution the 
interval Bellman equations the algorithm is solving coincide with the true Bellman equations 
of the environment, and thus $\underline{V}$ and $\overline{V}$ converge
pointwise towards the solution of the Bellman equations $V^{*}$ of the MDP $\cM$.

Lastly, as in the long run $\underline{V} = \overline{V} = V^{*}$, and
$\underline{\sigma}$ and $\overline{\sigma}$ maximize $\underline{V}$ and $\overline{V}$,
respectively, clearly both also maximize $V^{*}$ and thus converge towards optimal strategies.
\qed

\paragraph{}\noindent
Such random exploration schemes are also called \emph{dithering}. One well known disadvantage of such approaches
is that this unguided exploration does not differentiate between suboptimal actions \cite{SutBar18}.

Another viable sampling strategy would be to use the softmax action selection where, for each 
state-action pair $(s,a)$ we define a \emph{temperature} parameter $\tau_{s,a}$ 
and compute the strategy as

$$\sigma(s)(a) = \frac{e^{\underline{Q}(s,a)/\tau_{s,a}}}{\sum_{a'\in\Act(s)}e^{\underline{Q}(s,a')/\tau_{s,a}}}$$

Specifically, this allows of to perform exploration probabilistically, but unlike 
$\epsilon$-learning introduces a weight to distinguish between non-optimal actions.
One problem that arises however is that finding good parameters $\tau_{s,a}$ 
requires some degree of knowledge about possible action quality values.
In a resource constrained setting there is most likely not enough time to find 
suitable parameters, thus we focus on $\epsilon$-learning in the main part of the paper.
If however good estimates of action quality are present a priori, one might want
to consider other exploration schemes, such as the softmax sampling strategy.

\subsection{Runtime}

\begin{table}[]\caption{Model sizes and runtimes\label{tbl:runtimes}}
\resizebox{\columnwidth}{!}{
\begin{tabular}{l|c|ccc|cc}
               & \multicolumn{1}{l|}{}                 & \multicolumn{3}{c|}{model size}                 & \multicolumn{2}{c}{runtime {[}s{]}} \\
model          & \multicolumn{1}{l|}{optimal solution} & \multicolumn{1}{c|}{states} & \multicolumn{1}{l|}{state-action pairs} & transitions & \multicolumn{1}{c|}{no scoping} & \multicolumn{1}{c}{scoping} \\ \hline
Bandit      & 0.99                                  & \multicolumn{1}{c|}{3}    & \multicolumn{1}{c|}{99}                & 198         & \multicolumn{1}{c|}{0.17}     & \multicolumn{1}{c}{0.25}           \\
Bandit25-75      & 0.75                                  & \multicolumn{1}{c|}{3}    & \multicolumn{1}{c|}{101}                & 202         & \multicolumn{1}{c|}{0.17}     & \multicolumn{1}{c}{0.27}           \\
BanditGauss      & 0.7                                  & \multicolumn{1}{c|}{3}    & \multicolumn{1}{c|}{100}                & 200         & \multicolumn{1}{c|}{0.15}     & \multicolumn{1}{c}{0.24}           \\
BettingFav      & 17.5                                  & \multicolumn{1}{c|}{235}    & \multicolumn{1}{c|}{771}                & 1542         & \multicolumn{1}{c|}{2.2}     & \multicolumn{1}{c}{4.2}           \\
BettingUnfav      & 10                                  & \multicolumn{1}{c|}{235}    & \multicolumn{1}{c|}{771}                & 1542         & \multicolumn{1}{c|}{2.2}     & \multicolumn{1}{c}{3.2}           \\
consensus      & 0.11                                  & \multicolumn{1}{c|}{272}    & \multicolumn{1}{c|}{400}                & 492         & \multicolumn{1}{c|}{23}     & \multicolumn{1}{c}{34}           \\
csma           & 0.50                                  & \multicolumn{1}{c|}{1038}   & \multicolumn{1}{c|}{1054}               & 1282        & \multicolumn{1}{c|}{27}     & \multicolumn{1}{c}{30}             \\
Gridworld      & -5.5                                  & \multicolumn{1}{c|}{20}    & \multicolumn{1}{c|}{48}                & 132         & \multicolumn{1}{c|}{3.1}     & \multicolumn{1}{c}{5.2}           \\
pacman         & 0.45                                  & \multicolumn{1}{c|}{6854}   & \multicolumn{1}{c|}{8484}               & 8889        & \multicolumn{1}{c|}{7.6}     & \multicolumn{1}{c}{10}            \\
RacetrackSmall  & 0.49                                  & \multicolumn{1}{c|}{158}    & \multicolumn{1}{c|}{1377}               & 4608        & \multicolumn{1}{c|}{13}     & \multicolumn{1}{c}{19}          \\
RacetrackBig & 0.89                                  & \multicolumn{1}{c|}{12913}  & \multicolumn{1}{c|}{101701}             & 568334     & \multicolumn{1}{c|}{1874}     & \multicolumn{1}{c}{3326}           \\
SnL100         & -17.41                                & \multicolumn{1}{c|}{80}     & \multicolumn{1}{c|}{480}                & 1894        & \multicolumn{1}{c|}{99}  & \multicolumn{1}{c}{99}              \\
wlan\_cost      & -220                                  & \multicolumn{1}{c|}{2954}   & \multicolumn{1}{c|}{3972}               & 5202        & \multicolumn{1}{c|}{134}     & \multicolumn{1}{c}{22}            
\end{tabular}
}
\end{table}

In the main evaluation section we abstracted time constraints by
limiting each learning approach to 50 episodes with a fixed number of runs each.
In \Cref{tbl:runtimes} we give the real-time runtime for each algorithm
as well as the expected reward of the optimal strategy and model sizes.

The runtimes are given for the LCB samploing method, for both without scoping,
and with a scoping parameter $h=0.05$ and eager scoping.
Introducing scoping increases the runtime of up to a factor of 2, as it is
necessary to perform an additional value iteration to compute $\underline{V}_h$
and $\overline{V}_h$ as opposed to only $\underline{V}$ and $\overline{V}$.
We found that neither the sampling method nor the scoping method
(i.e., eager or conservative, and choice of $h$) have a measurable impact on the runtime.
The overhead from checking the scoping criterion is negligible.
In most examples,the reduction of the model does not yield a significantly faster value iteration
since its runtime is mostly dependant on the state-space whereas scopes
mostly only reduce the action space. However, in cases where a lot of actions
are removed from scopes, and certain states become unreachable, we do indeed
observe a significant speedup. This is most emphasized in the
\textrm{wlan\_cost} example.

In the table we can clearly see a correlation between model size and runtime.
The most contributing factors here are the state-space and the topological
complexity of the model, i.e., the number of transitions, especially in the
presence of loops which causes the value iteration to require more iterations.
Even though models with the reachability objective (those with optimal solution $\in [0,1])$)
can initialize the value interval to $[0,1]$ in each state
while models with cost-minimizing objectives (those with negative optimal solutions)
have to start with more conservative bounds, there is no clear correlation between
the reward structure and runtime.

\section{Further examples}
\begin{table}[t]

\caption{Optimistic and pessimistic strategy bounds for further examples\label{tbl:experiments}}
\resizebox{\columnwidth}{!}{%
\rotatebox{90}{
\begin{tabular}{lll|ccc|ccc}
\toprule
                           &              &                 & \multicolumn{3}{c|}{UCB}                             & \multicolumn{3}{c}{LCB}                              \\
\multicolumn{1}{l|}{model} & \multicolumn{1}{l|}{solution}  &   \multicolumn{1}{l|}{strategy}   & no scoping & conservative & eager & no scoping & conservative & eager \\ \midrule
\multicolumn{1}{l|}{\multirow{2}{*}{Bandit}} & \multicolumn{1}{l|}{\multirow{2}{*}{0.99}} & \multicolumn{1}{c|}{$\overline{\sigma}$}   & [0.71, 0.98] & [0.75, 0.98] & [0.88, 0.98] & [0.01, 0.99] & [0.04, 0.99] & [0.88, 0.98]  \\
\multicolumn{1}{l|}{} & \multicolumn{1}{l|}{} & \multicolumn{1}{c|}{$\underline{\sigma}$}  & [0.84, 0.98] & [0.86, 0.98] & [0.92, 0.98] & [0.87, 0.92] & [0.84, 0.89] & [0.92, 0.97] \\ \midrule
\multicolumn{1}{l|}{\multirow{2}{*}{Bandit25-75}} & \multicolumn{1}{l|}{\multirow{2}{*}{0.75}} & \multicolumn{1}{c|}{$\overline{\sigma}$}   & [0.28, 0.96] & [0.31, 0.95] & [0.71, 0.78] & [0.01, 0.99] & [0.01, 0.99] & [0.61, 0.90]  \\
\multicolumn{1}{l|}{} & \multicolumn{1}{l|}{} & \multicolumn{1}{c|}{$\underline{\sigma}$}  & [0.63, 0.92] & [0.64, 0.91] & [0.71, 0.78] & [0.61, 0.67] & [0.65, 0.72] & [0.69, 0.75] \\ \midrule
\multicolumn{1}{l|}{\multirow{2}{*}{BanditGauss}} & \multicolumn{1}{l|}{\multirow{2}{*}{0.7}} & \multicolumn{1}{c|}{$\overline{\sigma}$}   & [0.21, 0.94] & [0.18, 0.93] & [0.66, 0.72] & [0.01, 0.99] & [0.01, 0.99] & [0.48, 0.92]  \\
\multicolumn{1}{l|}{} & \multicolumn{1}{l|}{} & \multicolumn{1}{c|}{$\underline{\sigma}$}  & [0.57, 0.89] & [0.57, 0.87] & [0.66, 0.72] & [0.54, 0.60] & [0.57, 0.64] & [0.61, 0.68] \\ \midrule
\multicolumn{1}{l|}{\multirow{2}{*}{BettingFav}} & \multicolumn{1}{l|}{\multirow{2}{*}{17.55}} & \multicolumn{1}{c|}{$\overline{\sigma}$}   & [11.84, 23.67] & [12.13, 23.59] & [12.60, 22.40] & [0.12, 57.53] & [0.16, 57.09] & [0.54, 55.50]  \\
\multicolumn{1}{l|}{} & \multicolumn{1}{l|}{} & \multicolumn{1}{c|}{$\underline{\sigma}$}  & [13.17, 20.97] & [12.86, 22.01] & [13.18, 21.26] & [11.46, 12.54] & [11.73, 13.09] & [11.49, 13.03] \\ \midrule
\multicolumn{1}{l|}{\multirow{2}{*}{BettingUnfav}} & \multicolumn{1}{l|}{\multirow{2}{*}{10}} & \multicolumn{1}{c|}{$\overline{\sigma}$}   & [4.47, 11.63] & [4.79, 11.30] & [10.00, 10.00] & [0.00, 53.37] & [0.00, 53.70] & [0.39, 50.07]  \\
\multicolumn{1}{l|}{} & \multicolumn{1}{l|}{} & \multicolumn{1}{c|}{$\underline{\sigma}$}  & [10.00, 10.00] & [10.00, 10.00] & [10.00, 10.00] & [10.00, 10.00] & [10.00, 10.00] & [10.00, 10.00] \\ \midrule
\multicolumn{1}{l|}{\multirow{2}{*}{consensus}} & \multicolumn{1}{l|}{\multirow{2}{*}{0.1}} & \multicolumn{1}{c|}{$\overline{\sigma}$}   & [0.03, 0.25] & [0.04, 0.22] & [0.05, 0.19] & [0.00, 0.98] & [0.00, 0.90] & [0.01, 0.67]  \\
\multicolumn{1}{l|}{} & \multicolumn{1}{l|}{} & \multicolumn{1}{c|}{$\underline{\sigma}$}  & [0.03, 0.25] & [0.04, 0.22] & [0.05, 0.19] & [0.02, 0.14] & [0.03, 0.13] & [0.03, 0.12] \\ \midrule
\multicolumn{1}{l|}{\multirow{2}{*}{csma}} & \multicolumn{1}{l|}{\multirow{2}{*}{0.5}} & \multicolumn{1}{c|}{$\overline{\sigma}$}   & [0.44, 0.56] & [0.44, 0.56] & [0.44, 0.56] & [0.30, 0.69] & [0.32, 0.70] & [0.40, 0.61]  \\
\multicolumn{1}{l|}{} & \multicolumn{1}{l|}{} & \multicolumn{1}{c|}{$\underline{\sigma}$}  & [0.44, 0.56] & [0.44, 0.56] & [0.45, 0.56] & [0.46, 0.55] & [0.45, 0.55] & [0.46, 0.55] \\ \midrule
\multicolumn{1}{l|}{\multirow{2}{*}{Gridworld}} & \multicolumn{1}{l|}{\multirow{2}{*}{-5.48}} & \multicolumn{1}{c|}{$\overline{\sigma}$}   & [-6.26, -5.11] & [-6.23, -5.13] & [-6.21, -5.12] & [-67.53, -4.35] & [-65.07, -4.40] & [-45.05, -4.48]  \\
\multicolumn{1}{l|}{} & \multicolumn{1}{l|}{} & \multicolumn{1}{c|}{$\underline{\sigma}$}  & [-6.14, -5.13] & [-6.15, -5.13] & [-6.13, -5.13] & [-6.35, -5.17] & [-6.25, -5.15] & [-6.21, -5.11] \\ \midrule
\multicolumn{1}{l|}{\multirow{2}{*}{pacman}} & \multicolumn{1}{l|}{\multirow{2}{*}{0.45}} & \multicolumn{1}{c|}{$\overline{\sigma}$}   & [0.36, 0.53] & [0.37, 0.53] & [0.36, 0.53] & [0.37, 0.53] & [0.37, 0.53] & [0.36, 0.53]  \\
\multicolumn{1}{l|}{} & \multicolumn{1}{l|}{} & \multicolumn{1}{c|}{$\underline{\sigma}$}  & [0.36, 0.53] & [0.37, 0.53] & [0.36, 0.53] & [0.37, 0.53] & [0.37, 0.53] & [0.36, 0.53] \\ \midrule
\multicolumn{1}{l|}{\multirow{2}{*}{RacetrackBig}} & \multicolumn{1}{l|}{\multirow{2}{*}{0.89}} & \multicolumn{1}{c|}{$\overline{\sigma}$}   & [0.00, 0.99] & [0.00, 0.99] & [0.00, 1.00] & [0.00, 1.00] & [0.00, 1.00] & [0.00, 1.00]  \\
\multicolumn{1}{l|}{} & \multicolumn{1}{l|}{} & \multicolumn{1}{c|}{$\underline{\sigma}$}  & [0.28, 0.98] & [0.61, 0.95] & [0.51, 0.97] & [0.41, 0.87] & [0.43, 0.86] & [0.44, 0.86] \\ \midrule
\multicolumn{1}{l|}{\multirow{2}{*}{RacetrackSmall}} & \multicolumn{1}{l|}{\multirow{2}{*}{0.49}} & \multicolumn{1}{c|}{$\overline{\sigma}$}   & [0.11, 0.75] & [0.06, 0.75] & [0.18, 0.70] & [0.00, 0.95] & [0.00, 0.95] & [0.00, 0.95]  \\
\multicolumn{1}{l|}{} & \multicolumn{1}{l|}{} & \multicolumn{1}{c|}{$\underline{\sigma}$}  & [0.29, 0.74] & [0.30, 0.73] & [0.31, 0.69] & [0.33, 0.53] & [0.35, 0.52] & [0.34, 0.52] \\ \midrule
\multicolumn{1}{l|}{\multirow{2}{*}{SnL100}} & \multicolumn{1}{l|}{\multirow{2}{*}{-17.41}} & \multicolumn{1}{c|}{$\overline{\sigma}$}   & [-20.24, -15.96] & [-19.66, -15.83] & [-19.62, -15.86] & [-60.05, -13.60] & [-62.04, -13.72] & [-49.32, -13.83]  \\
\multicolumn{1}{l|}{} & \multicolumn{1}{l|}{} & \multicolumn{1}{c|}{$\underline{\sigma}$}  & [-19.41, -15.93] & [-19.42, -15.89] & [-19.38, -15.93] & [-20.03, -16.81] & [-19.84, -16.63] & [-19.86, -16.68] \\ \midrule
\multicolumn{1}{l|}{\multirow{2}{*}{wlan\_cost}} & \multicolumn{1}{l|}{\multirow{2}{*}{-220}} & \multicolumn{1}{c|}{$\overline{\sigma}$}   & [-226.27, -213.56] & [-225.00, -214.79] & [-224.99, -214.77] & [-226.49, -213.34] & [-224.98, -214.72] & [-225.03, -214.78]  \\
\multicolumn{1}{l|}{} & \multicolumn{1}{l|}{} & \multicolumn{1}{c|}{$\underline{\sigma}$}  & [-226.26, -213.58] & [-224.99, -214.80] & [-224.99, -214.77] & [-226.49, -213.34] & [-224.96, -214.75] & [-225.02, -214.78] \\
\bottomrule
\end{tabular}
}
}
\end{table}

Here we provide results for further examples of MDPs from the formal methods and RL community. 
Important model parameters as well as the value function  the optimistically optimal
strategy $\sigma$ and pessimistically optimal strategy $\sigma$ after the full 50 episodes for a
scoping parameter of $h=0.05$ are provided in \Cref{tbl:experiments}.
Additionally, we provide plots of the value function bounds for each example after each episode. 
For the experimental setup and a full description of the plotted values, we refer to \Cref{sec:evaluation}. 
\clearpage
    \subsection{Bandit}
    \begin{figure}[H]
 \begin{subfigure}[t]{.5\textwidth}
  \centering\scalebox{.8}{\input{img/appendix/BanditUCB_bounds}}
 \end{subfigure} \,
 \begin{subfigure}[t]{.5\textwidth}
  \centering\scalebox{.8}{\input{img/appendix/BanditLCB_bounds}}
 \end{subfigure}
\vspace{-1.5em}\caption{Bounds of the subsystem obtained by scoping (UCB left, LCB right)\vspace{1em}\label{fig:habit_bounds_Bandit}}
 \begin{subfigure}[t]{.5\textwidth}
  \centering\scalebox{.8}{\input{img/appendix/BanditUCB_corr_bounds}}
 \end{subfigure} \,
 \begin{subfigure}[t]{.5\textwidth}
  \centering\scalebox{.8}{\input{img/appendix/BanditLCB_corr_bounds}}
 \end{subfigure}
\vspace{-1.5em}\caption{Bounds for pessimistically optimal strategy (UCB left, LCB right)\vspace{1em}\label{fig:habit_corr_bounds_Bandit}}

 \begin{subfigure}[t]{.5\textwidth}
  \centering\scalebox{.8}{\input{img/appendix/BanditUCB_real}}
 \end{subfigure} \,
 \begin{subfigure}[t]{.5\textwidth}
  \centering\scalebox{.8}{\input{img/appendix/BanditLCB_real}}
 \end{subfigure}
\vspace{-1.5em}\caption{Expected total reward in $\cM$ w.r.t. policy $\sigma$ (UCB left, LCB right)\label{fig:real_strategy_Bandit}}

\end{figure}

    \subsection{Bandit25-75}
    \begin{figure}[H]
 \begin{subfigure}[t]{.5\textwidth}
  \centering\scalebox{.8}{\input{img/appendix/Bandit25-75UCB_bounds}}
 \end{subfigure} \,
 \begin{subfigure}[t]{.5\textwidth}
  \centering\scalebox{.8}{\input{img/appendix/Bandit25-75LCB_bounds}}
 \end{subfigure}
\vspace{-1.5em}\caption{Bounds of the subsystem obtained by scoping (UCB left, LCB right)\vspace{1em}
\label{fig:habit_bounds_Bandit25-75_appendix}}
 \begin{subfigure}[t]{.5\textwidth}
  \centering\scalebox{.8}{\input{img/appendix/Bandit25-75UCB_corr_bounds}}
 \end{subfigure} \,
 \begin{subfigure}[t]{.5\textwidth}
  \centering\scalebox{.8}{\input{img/appendix/Bandit25-75LCB_corr_bounds}}
 \end{subfigure}
\vspace{-1.5em}\caption{Bounds for pessimistically optimal strategy (UCB left, LCB right)\vspace{1em}\label{fig:habit_corr_bounds_Bandit25-75_appendix}}

 \begin{subfigure}[t]{.5\textwidth}
  \centering\scalebox{.8}{\input{img/appendix/Bandit25-75UCB_real}}
 \end{subfigure} \,
 \begin{subfigure}[t]{.5\textwidth}
  \centering\scalebox{.8}{\input{img/appendix/Bandit25-75LCB_real}}
 \end{subfigure}
\vspace{-1.5em}\caption{Expected total reward in $\cM$ w.r.t. policy $\sigma$ (UCB left, LCB right)\label{fig:real_strategy_Bandit25-75_appendix}}

\end{figure}

    \subsection{BanditGauss}
    \begin{figure}[H]
 \begin{subfigure}[t]{.5\textwidth}
  \centering\scalebox{.8}{\input{img/appendix/BanditGaussUCB_bounds}}
 \end{subfigure} \,
 \begin{subfigure}[t]{.5\textwidth}
  \centering\scalebox{.8}{\input{img/appendix/BanditGaussLCB_bounds}}
 \end{subfigure}
\vspace{-1.5em}\caption{Bounds of the subsystem obtained by scoping (UCB left, LCB right)\vspace{1em}
\label{fig:habit_bounds_BanditGauss}}
 \begin{subfigure}[t]{.5\textwidth}
  \centering\scalebox{.8}{\input{img/appendix/BanditGaussUCB_corr_bounds}}
 \end{subfigure} \,
 \begin{subfigure}[t]{.5\textwidth}
  \centering\scalebox{.8}{\input{img/appendix/BanditGaussLCB_corr_bounds}}
 \end{subfigure}
\vspace{-1.5em}\caption{Bounds for pessimistically optimal strategy (UCB left, LCB right)\vspace{1em}\label{fig:habit_corr_bounds_BanditGauss}}

 \begin{subfigure}[t]{.5\textwidth}
  \centering\scalebox{.8}{\input{img/appendix/BanditGaussUCB_real}}
 \end{subfigure} \,
 \begin{subfigure}[t]{.5\textwidth}
  \centering\scalebox{.8}{\input{img/appendix/BanditGaussLCB_real}}
 \end{subfigure}
\vspace{-1.5em}\caption{Expected total reward in $\cM$ w.r.t. policy $\sigma$ (UCB left, LCB right)\label{fig:real_strategy_BanditGauss}}

\end{figure}

    \subsection{BettingFav}
    \begin{figure}[H]
 \begin{subfigure}[t]{.5\textwidth}
  \centering\scalebox{.8}{\input{img/appendix/BettingFavUCB_bounds}}
 \end{subfigure} \,
 \begin{subfigure}[t]{.5\textwidth}
  \centering\scalebox{.8}{\input{img/appendix/BettingFavLCB_bounds}}
 \end{subfigure}
\vspace{-1.5em}\caption{Bounds of the subsystem obtained by scoping (UCB left, LCB right)\vspace{1em}
\label{fig:habit_bounds_BettingFav}}
 \begin{subfigure}[t]{.5\textwidth}
  \centering\scalebox{.8}{\input{img/appendix/BettingFavUCB_corr_bounds}}
 \end{subfigure} \,
 \begin{subfigure}[t]{.5\textwidth}
  \centering\scalebox{.8}{\input{img/appendix/BettingFavLCB_corr_bounds}}
 \end{subfigure}
\vspace{-1.5em}\caption{Bounds for pessimistically optimal strategy (UCB left, LCB right)\vspace{1em}\label{fig:habit_corr_bounds_BettingFav}}

 \begin{subfigure}[t]{.5\textwidth}
  \centering\scalebox{.8}{\input{img/appendix/BettingFavUCB_real}}
 \end{subfigure} \,
 \begin{subfigure}[t]{.5\textwidth}
  \centering\scalebox{.8}{\input{img/appendix/BettingFavLCB_real}}
 \end{subfigure}
\vspace{-1.5em}\caption{Expected total reward in $\cM$ w.r.t. policy $\sigma$ (UCB left, LCB right)\label{fig:real_strategy_BettingFav}}

\end{figure}

    \subsection{BettingUnfav}
    \begin{figure}[H]
 \begin{subfigure}[t]{.5\textwidth}
  \centering\scalebox{.8}{\input{img/appendix/BettingUnfavUCB_bounds}}
 \end{subfigure} \,
 \begin{subfigure}[t]{.5\textwidth}
  \centering\scalebox{.8}{\input{img/appendix/BettingUnfavLCB_bounds}}
 \end{subfigure}
\vspace{-1.5em}\caption{Bounds of the subsystem obtained by scoping (UCB left, LCB right)\vspace{1em}
\label{fig:habit_bounds_BettingUnfav}}
 \begin{subfigure}[t]{.5\textwidth}
  \centering\scalebox{.8}{\input{img/appendix/BettingUnfavUCB_corr_bounds}}
 \end{subfigure} \,
 \begin{subfigure}[t]{.5\textwidth}
  \centering\scalebox{.8}{\input{img/appendix/BettingUnfavLCB_corr_bounds}}
 \end{subfigure}
\vspace{-1.5em}\caption{Bounds for pessimistically optimal strategy (UCB left, LCB right)\vspace{1em}\label{fig:habit_corr_bounds_BettingUnfav}}

 \begin{subfigure}[t]{.5\textwidth}
  \centering\scalebox{.8}{\input{img/appendix/BettingUnfavUCB_real}}
 \end{subfigure} \,
 \begin{subfigure}[t]{.5\textwidth}
  \centering\scalebox{.8}{\input{img/appendix/BettingUnfavLCB_real}}
 \end{subfigure}
\vspace{-1.5em}\caption{Expected total reward in $\cM$ w.r.t. policy $\sigma$ (UCB left, LCB right)\label{fig:real_strategy_BettingUnfav}}

\end{figure}

    \subsection{consensus}
    \begin{figure}[H]
 \begin{subfigure}[t]{.5\textwidth}
  \centering\scalebox{.8}{\input{img/appendix/consensusUCB_bounds}}
 \end{subfigure} \,
 \begin{subfigure}[t]{.5\textwidth}
  \centering\scalebox{.8}{\input{img/appendix/consensusLCB_bounds}}
 \end{subfigure}
\vspace{-1.5em}\caption{Bounds of the subsystem obtained by scoping (UCB left, LCB right)\vspace{1em}
\label{fig:habit_bounds_consensus}}
 \begin{subfigure}[t]{.5\textwidth}
  \centering\scalebox{.8}{\input{img/appendix/consensusUCB_corr_bounds}}
 \end{subfigure} \,
 \begin{subfigure}[t]{.5\textwidth}
  \centering\scalebox{.8}{\input{img/appendix/consensusLCB_corr_bounds}}
 \end{subfigure}
\vspace{-1.5em}\caption{Bounds for pessimistically optimal strategy (UCB left, LCB right)\vspace{1em}\label{fig:habit_corr_bounds_consensus}}

 \begin{subfigure}[t]{.5\textwidth}
  \centering\scalebox{.8}{\input{img/appendix/consensusUCB_real}}
 \end{subfigure} \,
 \begin{subfigure}[t]{.5\textwidth}
  \centering\scalebox{.8}{\input{img/appendix/consensusLCB_real}}
 \end{subfigure}
\vspace{-1.5em}\caption{Expected total reward in $\cM$ w.r.t. policy $\sigma$ (UCB left, LCB right)\label{fig:real_strategy_consensus}}

\end{figure}

    \subsection{csma}
    \begin{figure}[H]
 \begin{subfigure}[t]{.5\textwidth}
  \centering\scalebox{.8}{\input{img/appendix/csmaUCB_bounds}}
 \end{subfigure} \,
 \begin{subfigure}[t]{.5\textwidth}
  \centering\scalebox{.8}{\input{img/appendix/csmaLCB_bounds}}
 \end{subfigure}
\vspace{-1.5em}\caption{Bounds of the subsystem obtained by scoping (UCB left, LCB right)\vspace{1em}
\label{fig:habit_bounds_csma}}
 \begin{subfigure}[t]{.5\textwidth}
  \centering\scalebox{.8}{\input{img/appendix/csmaUCB_corr_bounds}}
 \end{subfigure} \,
 \begin{subfigure}[t]{.5\textwidth}
  \centering\scalebox{.8}{\input{img/appendix/csmaLCB_corr_bounds}}
 \end{subfigure}
\vspace{-1.5em}\caption{Bounds for pessimistically optimal strategy (UCB left, LCB right)\vspace{1em}\label{fig:habit_corr_bounds_csma}}

 \begin{subfigure}[t]{.5\textwidth}
  \centering\scalebox{.8}{\input{img/appendix/csmaUCB_real}}
 \end{subfigure} \,
 \begin{subfigure}[t]{.5\textwidth}
  \centering\scalebox{.8}{\input{img/appendix/csmaLCB_real}}
 \end{subfigure}
\vspace{-1.5em}\caption{Expected total reward in $\cM$ w.r.t. policy $\sigma$ (UCB left, LCB right)\label{fig:real_strategy_csma}}

\end{figure}

    \subsection{Gridworld}
    \begin{figure}[H]
 \begin{subfigure}[t]{.5\textwidth}
  \centering\scalebox{.8}{\input{img/appendix/GridworldUCB_bounds}}
 \end{subfigure} \,
 \begin{subfigure}[t]{.5\textwidth}
  \centering\scalebox{.8}{\input{img/appendix/GridworldLCB_bounds}}
 \end{subfigure}
\vspace{-1.5em}\caption{Bounds of the subsystem obtained by scoping (UCB left, LCB right)\vspace{1em}
\label{fig:habit_bounds_Gridworld}}
 \begin{subfigure}[t]{.5\textwidth}
  \centering\scalebox{.8}{\input{img/appendix/GridworldUCB_corr_bounds}}
 \end{subfigure} \,
 \begin{subfigure}[t]{.5\textwidth}
  \centering\scalebox{.8}{\input{img/appendix/GridworldLCB_corr_bounds}}
 \end{subfigure}
\vspace{-1.5em}\caption{Bounds for pessimistically optimal strategy (UCB left, LCB right)\vspace{1em}\label{fig:habit_corr_bounds_Gridworld}}

 \begin{subfigure}[t]{.5\textwidth}
  \centering\scalebox{.8}{\input{img/appendix/GridworldUCB_real}}
 \end{subfigure} \,
 \begin{subfigure}[t]{.5\textwidth}
  \centering\scalebox{.8}{\input{img/appendix/GridworldLCB_real}}
 \end{subfigure}
\vspace{-1.5em}\caption{Expected total reward in $\cM$ w.r.t. policy $\sigma$ (UCB left, LCB right)\label{fig:real_strategy_Gridworld}}

\end{figure}

    \subsection{pacman}
    \begin{figure}[H]
 \begin{subfigure}[t]{.5\textwidth}
  \centering\scalebox{.8}{\input{img/appendix/pacmanUCB_bounds}}
 \end{subfigure} \,
 \begin{subfigure}[t]{.5\textwidth}
  \centering\scalebox{.8}{\input{img/appendix/pacmanLCB_bounds}}
 \end{subfigure}
\vspace{-1.5em}\caption{Bounds of the subsystem obtained by scoping (UCB left, LCB right)\vspace{1em}
\label{fig:habit_bounds_pacman}}
 \begin{subfigure}[t]{.5\textwidth}
  \centering\scalebox{.8}{\input{img/appendix/pacmanUCB_corr_bounds}}
 \end{subfigure} \,
 \begin{subfigure}[t]{.5\textwidth}
  \centering\scalebox{.8}{\input{img/appendix/pacmanLCB_corr_bounds}}
 \end{subfigure}
\vspace{-1.5em}\caption{Bounds for pessimistically optimal strategy (UCB left, LCB right)\vspace{1em}\label{fig:habit_corr_bounds_pacman}}

 \begin{subfigure}[t]{.5\textwidth}
  \centering\scalebox{.8}{\input{img/appendix/pacmanUCB_real}}
 \end{subfigure} \,
 \begin{subfigure}[t]{.5\textwidth}
  \centering\scalebox{.8}{\input{img/appendix/pacmanLCB_real}}
 \end{subfigure}
\vspace{-1.5em}\caption{Expected total reward in $\cM$ w.r.t. policy $\sigma$ (UCB left, LCB right)\label{fig:real_strategy_pacman}}

\end{figure}

    \subsection{RacetrackBig}
    \begin{figure}[H]
 \begin{subfigure}[t]{.5\textwidth}
  \centering\scalebox{.8}{\input{img/appendix/RacetrackSmallUCB_bounds}}
 \end{subfigure} \,
 \begin{subfigure}[t]{.5\textwidth}
  \centering\scalebox{.8}{\input{img/appendix/RacetrackSmallLCB_bounds}}
 \end{subfigure}
\vspace{-1.5em}\caption{Bounds of the subsystem obtained by scoping (UCB left, LCB right)\vspace{1em}\label{fig:habit_bounds_RacetrackBig}}
 \begin{subfigure}[t]{.5\textwidth}
  \centering\scalebox{.8}{\input{img/appendix/RacetrackSmallUCB_corr_bounds}}
 \end{subfigure} \,
 \begin{subfigure}[t]{.5\textwidth}
  \centering\scalebox{.8}{\input{img/appendix/RacetrackSmallLCB_corr_bounds}}
 \end{subfigure}
\vspace{-1.5em}\caption{Bounds for pessimistically optimal strategy (UCB left, LCB right)\vspace{1em}\label{fig:habit_corr_bounds_RacetrackBig}}

 \begin{subfigure}[t]{.5\textwidth}
  \centering\scalebox{.8}{\input{img/appendix/RacetrackSmallUCB_real}}
 \end{subfigure} \,
 \begin{subfigure}[t]{.5\textwidth}
  \centering\scalebox{.8}{\input{img/appendix/RacetrackSmallLCB_real}}
 \end{subfigure}
\vspace{-1.5em}\caption{Expected total reward in $\cM$ w.r.t. policy $\sigma$ (UCB left, LCB right)\label{fig:real_strategy_RacetrackBig}}

\end{figure}

    \subsection{RacetrackSmall}
    \begin{figure}[H]
 \begin{subfigure}[t]{.5\textwidth}
  \centering\scalebox{.8}{\input{img/appendix/RacetrackTinyUCB_bounds}}
 \end{subfigure} \,
 \begin{subfigure}[t]{.5\textwidth}
  \centering\scalebox{.8}{\input{img/appendix/RacetrackTinyLCB_bounds}}
 \end{subfigure}
\vspace{-1.5em}\caption{Bounds of the subsystem obtained by scoping (UCB left, LCB right)\vspace{1em}\label{fig:habit_bounds_RacetrackSmall}}
 \begin{subfigure}[t]{.5\textwidth}
  \centering\scalebox{.8}{\input{img/appendix/RacetrackTinyUCB_corr_bounds}}
 \end{subfigure} \,
 \begin{subfigure}[t]{.5\textwidth}
  \centering\scalebox{.8}{\input{img/appendix/RacetrackTinyLCB_corr_bounds}}
 \end{subfigure}
\vspace{-1.5em}\caption{Bounds for pessimistically optimal strategy (UCB left, LCB right)\vspace{1em}\label{fig:habit_corr_bounds_RacetrackSmall}}

 \begin{subfigure}[t]{.5\textwidth}
  \centering\scalebox{.8}{\input{img/appendix/RacetrackTinyUCB_real}}
 \end{subfigure} \,
 \begin{subfigure}[t]{.5\textwidth}
  \centering\scalebox{.8}{\input{img/appendix/RacetrackTinyLCB_real}}
 \end{subfigure}
\vspace{-1.5em}\caption{Expected total reward in $\cM$ w.r.t. policy $\sigma$ (UCB left, LCB right)\label{fig:real_strategy_RacetrackSmall}}

\end{figure}

    \subsection{SnL100}
    \begin{figure}[H]
 \begin{subfigure}[t]{.5\textwidth}
  \centering\scalebox{.8}{\input{img/appendix/SnL100UCB_bounds}}
 \end{subfigure} \,
 \begin{subfigure}[t]{.5\textwidth}
  \centering\scalebox{.8}{\input{img/appendix/SnL100LCB_bounds}}
 \end{subfigure}
\vspace{-1.5em}\caption{Bounds of the subsystem obtained by scoping (UCB left, LCB right)\vspace{1em}\label{fig:habit_bounds_SnL100}}
 \begin{subfigure}[t]{.5\textwidth}
  \centering\scalebox{.8}{\input{img/appendix/SnL100UCB_corr_bounds}}
 \end{subfigure} \,
 \begin{subfigure}[t]{.5\textwidth}
  \centering\scalebox{.8}{\input{img/appendix/SnL100LCB_corr_bounds}}
 \end{subfigure}
\vspace{-1.5em}\caption{Bounds for pessimistically optimal strategy (UCB left, LCB right)\vspace{1em}\label{fig:habit_corr_bounds_SnL100}}

 \begin{subfigure}[t]{.5\textwidth}
  \centering\scalebox{.8}{\input{img/appendix/SnL100UCB_real}}
 \end{subfigure} \,
 \begin{subfigure}[t]{.5\textwidth}
  \centering\scalebox{.8}{\input{img/appendix/SnL100LCB_real}}
 \end{subfigure}
\vspace{-1.5em}\caption{Expected total reward in $\cM$ w.r.t. policy $\sigma$ (UCB left, LCB right)\label{fig:real_strategy_SnL100}}

\end{figure}

    \subsection{wlan\_cost}
    \begin{figure}[H]
 \begin{subfigure}[t]{.5\textwidth}
  \centering\scalebox{.8}{\input{img/appendix/wlan_costUCB_bounds}}
 \end{subfigure} \,
 \begin{subfigure}[t]{.5\textwidth}
  \centering\scalebox{.8}{\input{img/appendix/wlan_costLCB_bounds}}
 \end{subfigure}
\vspace{-1.5em}\caption{Bounds of the subsystem obtained by scoping (UCB left, LCB right)\vspace{1em}\label{fig:habit_bounds_wlan_cost}}
 \begin{subfigure}[t]{.5\textwidth}
  \centering\scalebox{.8}{\input{img/appendix/wlan_costUCB_corr_bounds}}
 \end{subfigure} \,
 \begin{subfigure}[t]{.5\textwidth}
  \centering\scalebox{.8}{\input{img/appendix/wlan_costLCB_corr_bounds}}
 \end{subfigure}
\vspace{-1.5em}\caption{Bounds for pessimistically optimal strategy (UCB left, LCB right)\vspace{1em}\label{fig:habit_corr_bounds_wlan_cost}}

 \begin{subfigure}[t]{.5\textwidth}
  \centering\scalebox{.8}{\input{img/appendix/wlan_costUCB_real}}
 \end{subfigure} \,
 \begin{subfigure}[t]{.5\textwidth}
  \centering\scalebox{.8}{\input{img/appendix/wlan_costLCB_real}}
 \end{subfigure}
\vspace{-1.5em}\caption{Expected total reward in $\cM$ w.r.t. policy $\sigma$ (UCB left, LCB right)\label{fig:real_strategy_wlan_cost}}

\end{figure}

\end{document}